\documentclass[dvipsnames,table]{article} 
\usepackage{acl2020,times}

\usepackage{amsmath,amsfonts,bm}

\def\eqref#1{equation~\ref{#1}}

\def\1{\bm{1}}

\def\mM{{\bm{M}}}

\DeclareMathAlphabet{\mathsfit}{\encodingdefault}{\sfdefault}{m}{sl}
\SetMathAlphabet{\mathsfit}{bold}{\encodingdefault}{\sfdefault}{bx}{n}

\def\gX{{\mathcal{X}}}

\newcommand{\E}{\mathbb{E}}

\DeclareMathOperator*{\argmin}{arg\,min}

\usepackage{hyperref}
\usepackage{url}

\usepackage{latexsym}
\usepackage{soul}

\usepackage{boxedminipage}
\usepackage{xcolor}
\usepackage{bm}
\usepackage{xspace}
\usepackage{enumitem}
\usepackage{cuted}
\usepackage{wrapfig}
\usepackage[normalem]{ulem}

\usepackage{microtype}
\usepackage{multirow}
\usepackage{booktabs}

\usepackage{tikz}
\usepackage{pgfplots}
\pgfplotsset{compat=1.9}
\usetikzlibrary{patterns}
\usepackage{txfonts}

\newcommand{\az}{\textsc{Az}}
\newcommand{\be}{\textsc{Be}}
\newcommand{\cs}{\textsc{Cs}}
\newcommand{\en}{\textsc{En}}
\newcommand{\es}{\textsc{Es}}
\newcommand{\gl}{\textsc{Gl}}
\newcommand{\pt}{\textsc{Pt}}
\newcommand{\ru}{\textsc{Ru}}
\newcommand{\sk}{\textsc{Sk}}
\newcommand{\tr}{\textsc{Tr}}
\newcommand{\fr}{\textsc{Fr}}
\newcommand{\hi}{\textsc{Hi}}
\newcommand{\ko}{\textsc{Ko}}
\newcommand{\sv}{\textsc{Sv}}
\newcommand{\uk}{\textsc{Uk}}

\newcommand{\norm}[1]{\left\Vert #1 \right\Vert}

\usepackage[LGR,T1]{fontenc}
\usepackage[utf8]{inputenc}
\usepackage[greek,english]{babel}

\title{Should All Cross-Lingual Embeddings Speak English?}

\author{Antonios Anastasopoulos and Graham Neubig\\
Language Technologies Institute, Carnegie Mellon University\\
\texttt{\{aanastas,gneubig\}@cs.cmu.edu}}
\date{}
\aclfinalcopy
\begin{document}
\maketitle
\begin{abstract}
Most of recent work in cross-lingual word embeddings is severely Anglocentric. The vast majority of lexicon induction evaluation dictionaries are between English and another language, and the English embedding space is selected by default as the \textit{hub} when learning in a multilingual setting. With this work, however, we challenge these practices. First, we show that the choice of hub language can significantly impact downstream lexicon induction and zero-shot POS tagging performance. Second, we both expand a standard English-centered evaluation dictionary collection to include all language pairs using triangulation, and create new dictionaries for under-represented languages.\footnote{Available at \url{https://github.com/antonisa/embeddings}.} 
Evaluating established methods over all these language pairs sheds light into their suitability for aligning embeddings from distant languages and presents new challenges for the field.
Finally, in our analysis we identify general guidelines for strong cross-lingual embedding baselines, that extend to language pairs that do not include English. 
\end{abstract}

\section{Introduction}

Continuous vectors for representing words (embeddings) \citep{turian-etal-2010-word} have become ubiquitous in modern, neural NLP. Cross-lingual representations \citep{mikolov2013exploiting} additionally represent words from various languages in a shared continuous space, which in turn can be used for Bilingual Lexicon Induction (BLI).
BLI is often the first step towards several downstream tasks such as Part-Of-Speech (POS) tagging \citep{zhang2016ten}, parsing \citep{ammar2016many}, document classification \citep{klementiev2012inducing}, and machine translation \citep{irvine2013combining,artetxe-etal-2018-unsupervised,lample-etal-2018-phrase}.

Often, such shared representations are learned with a two-step process, whether under bilingual or multilingual settings (hereinafter BWE and MWE, respectively). First, monolingual word embeddings are learned over large swaths of text.
Such pre-trained word embeddings, such as the fastText Wikipedia vectors~\citep{grave2018learning}, are available for many languages and are widely used.
Second, a mapping between the languages is learned in one of three ways: in a supervised manner if dictionaries or parallel data are available to be used for supervision \citep{zou2013bilingual}, under minimal supervision e.g.~using only identical strings \citep{smith2017offline}, or even in an unsupervised fashion \citep{zhang2017adversarial,conneau2018word}.
Both in bilingual and multilingual settings, it is common that one of the language embedding spaces is the target to which all other languages get aligned (hereinafter ``the \textit{hub}"). We outline the details in Section~\ref{sec:mwe}.

Despite all the recent progress in learning cross-lingual embeddings, we identify a major shortcoming to previous work: it is by and large English-centric. 
Notably, most MWE approaches essentially select English as the \textit{hub} during training by default, aligning all other language spaces to the English one.
We argue and empirically show, however, that English is a poor hub language choice.
In BWE settings, on the other hand, it is fairly uncommon to denote which of the two languages is the \textit{hub} (often this is implied to be the target language). However, we experimentally find that this choice can greatly impact downstream performance, especially when aligning distant languages.

This Anglocentricity is even more evident at the evaluation stage. The lexica most commonly used for evaluation are the MUSE lexica \citep{conneau2018word} which cover 45 languages, but with translations only from and into English. Alternative evaluation dictionaries are also very English- and European-centric: \cite{dinu2014make} report results on English--Italian, \cite{artetxe-etal-2017-learning} on English--German and English--Finnish, \cite{zhang2017adversarial} on Spanish--English and Italian--English, and \cite{artetxe2018robust} between English and Italian, German, Finnish, Spanish, and Turkish.
We argue that cross-lingual word embedding mapping methods should look beyond English for their evaluation benchmarks because, compared to all others, English is a language with disproportionately large available data and relatively poor inflectional morphology e.g., it lacks case, gender, and complex verbal inflection systems~\citep{aronoff2011morphology}. These two factors allow for an overly easy evaluation setting which does not necessarily generalize to other language pairs.
In light of this, equal focus should instead be devoted to evaluation over more diverse language pairs that also include morphologically rich and low-resource languages.

With this work, we attempt to address these shortcomings, providing the following contributions:
\begin{itemize}
    \item We show that the choice of the \textit{hub} when evaluating on diverse language pairs can lead to significantly different performance for iterative refinement methods that use a symbolic-based seed dictionary (e.g., by more than~10 percentage points for BWE over distant languages).
    We also show that often English is a suboptimal hub for~MWE.
    \item We identify some general guidelines for choosing a hub language which could lead to stronger performance; \textit{less} isometry between the hub and source and target embedding spaces mildly correlates with performance, as does typological distance (a measure of language similarity based on language family membership trees). For distant languages, multilingual systems should be preferred over bilingual ones if the languages share alphabets, otherwise a bilingual system based on monolingual similarity dictionaries is preferable.
    \item We provide resources for training and evaluation on language pairs that do not include English. We outline a simple triangulation method with which we extend the MUSE dictionaries to an additional~4704 lexicons covering 50 languages (for a total of 4900 dictionaries, including the original English ones), and we present results on a subset of them. We also create new evaluation lexica for under-resourced, under-represented languages using Azerbaijani, Belarusian, and Galician as our test cases. Finally, we provide recipes for creating such dictionaries for any language pair with available parallel data.
\end{itemize}

\section{Cross-Lingual Word Embeddings and Lexicon Induction}
\label{sec:mwe}

\paragraph{Bilingual Word Embeddings}
In the supervised BWE setting of \citet{mikolov2013exploiting}, given two languages $\mathcal{L} = \{l_1,l_2\}$ and their pre-trained \textit{row-aligned} embeddings $\mathcal{X}_1, \mathcal{X}_2,$ respectively, a transformation matrix $\mM$ is learned such that: 
\[
    \mM = \argmin_{\mM \in \Omega} \norm{\gX_1 - \mM\gX_2}.
\]
The set $\Omega$ can potentially impose a constraint over $\mM$, such as the very popular constraint of restricting it to be orthogonal \citep{xing-etal-2015-normalized}.
Previous work has empirically found that this simple formulation is competitive with other more complicated alternatives \citep{xing-etal-2015-normalized}. The orthogonality assumption ensures that there exists a closed-form solution through Singular Value Decomposition (SVD) of $\gX_1\gX_2^T$.\footnote{We refer the reader to \citet{mikolov2013exploiting} for details.}
Note that in this case only a single matrix $\mM$ needs to be learned, because $\norm{\gX_1 - \mM^{}\gX_2}=\norm{\mM^{-1}\gX_1 - \gX_2}$, while at the same time a model that minimizes $\norm{\gX_1 - \mM\gX_2}$ is as expressive as one minimizing $\norm{\mM_1\gX_1 - \mM_2\gX_2}$, with half the parameters.

In the minimally supervised or even the unsupervised setting, \citet{zhang2017adversarial} and \citet{conneau2018word} reframe the task as an adversarial game, with a generator aiming to produce a transformation that can fool a discriminator. However, the most popular methods follow an iterative refinement approach~\citep{artetxe-etal-2017-learning}. Starting with a seed dictionary (e.g. from identical strings~\citep{zhou19naacl} or numerals) an initial mapping is learned in the same manner as in the supervised setting. The initial mapping, in turn, is used to expand the seed dictionary with high confidence word translation pairs. The new dictionary is then used to learn a better mapping, and so forth the iterations continue until convergence.
The same iterative approach is followed by \citet{artetxe2018robust}, with one important difference that allows their model (\texttt{VecMap}) to handle language pairs with different alphabets: instead of identical strings, the seed dictionary is constructed based on the similarity of the monolingual similarity distributions over all words in the vocabulary.\footnote{We refer the reader to~\citet{artetxe2018robust} for details.}



\paragraph{Multilingual Word Embeddings}
In a multilingual setting, the simplest approach is to use BWE and align all languages into a target language (the \textit{hub}). In this case, for $N$ languages $\mathcal{L} = \{l_1,l_2,\ldots,l_N\}$ on has to learn $N-1$ bilingual mappings \citep{ammar2016massively}.
Rather than using a single hub space, \citet{heyman-etal-2019-learning} propose an incremental procedure that uses an Incremental Hub Space (\texttt{IHS}): each new language is included to the multilingual space by mapping it to all languages that have already been aligned (e.g. language $l_3$ would be mapped to the aligned space of $\{l_1,l_2\}$).

Alternatively, all mappings could be learned jointly, taking advantage of the inter-dependencies between any two language pairs.
Importantly, though, there is no closed form solution for learning the joint mapping, hence a solution needs to be approximated with gradient-based methods.
The main approaches are:
\begin{itemize}
    \item Multilingual adversarial training with pseudo-randomized refinement~\cite[\texttt{MAT+MPSR}]{chen-cardie-2018-unsupervised}: a generalization of the adversarial approach of \citet{zhang2017adversarial,conneau2018word} to multiple languages, also combined with an iterative refinement procedure.\footnote{\texttt{MAT+MPSR} has the beneficial property of being as computationally efficient as learning $\mathcal{O}(N)$ mappings (instead of $\mathcal{O}(N^2)$). We refer the reader to \citet{chen-cardie-2018-unsupervised} for exact details.}
    \item Unsupervised Multilingual Hyperalignment~\cite[\texttt{UMH}]{alaux2018unsupervised}: an approach which maps all languages to a single \textit{hub} space,\footnote{Note that \citet{alaux2018unsupervised} use the term \textit{pivot} to refer to what we refer to as the \textit{hub} language.} but also enforces good alignments between all language pairs within this space.
\end{itemize}
Even though the architecture and modeling approach of all MWE methods are different, they share the same conceptual traits: one of the language spaces remains invariant and all other languages are effectively mapped to it. 
In all cases, English is by default selected to be the hub. The only exception is the study of triplets alignments in \cite{alaux2018unsupervised}, where Spanish is used as the Spanish--French--Portuguese triplet hub. 


\paragraph{Lexicon Induction} One of the most common downstream evaluation tasks for the learned cross-lingual word mappings is Lexicon Induction (LI), the task of retrieving the most appropriate word-level translation for a query word from the mapped embedding spaces. Specialized evaluation (and training) dictionaries have been created for multiple language pairs.
Of these, the MUSE dictionaries \citep{conneau2018word} are most often used, providing word translations between English (\en) and~48 other high- to mid-resource languages, as well as on all~30 pairs among~6 very similar Romance and Germanic languages (English, French, German, Spanish, Italian, Portuguese).

Given the mapped embedding spaces, the translations are retrieved using a distance metric, with Cross-Lingual Similarity Scaling \citep[CSLS]{conneau2018word} as the most commonly used in the literature. Intuitively, CSLS decreases the scores of pairs that lie in dense areas, increasing the scores of rarer words (which are harder to align).
The retrieved pairs are compared to the gold standard and evaluated using precision at $k$ (P@$k$, evaluating how often the correct translation is within the $k$ retrieved nearest neighbors of the query). Throughout this work we report P@1, which is equivalent to accuracy; we provide P@5 and P@10 results in the Appendix.

\section{New LI Evaluation Dictionaries}
\begin{table*}[t]
    \centering
    \begin{tabular}{cc|cc||ccc}
    \toprule
        \multicolumn{2}{c|}{Greek} & \multicolumn{2}{c||}{Italian} & \multicolumn{3}{c}{Bridged Greek--Italian Lexicon} \\
        word & tag & word & tag & Match & Greek & Italian \\
    \midrule
    {\foreignlanguage{greek}{ειρηνικός}} & {\small\texttt{M;NOM;SG}} & pacifico & {\small\texttt{M;SG}} & {\small\texttt{M;SG}} & {\foreignlanguage{greek}{ειρηνικός}} & pacifico, \st{pacifici}, pacifica\\
    {\foreignlanguage{greek}{ειρηνική}} & {\small\texttt{F;NOM;SG}} & pacifici & {\small\texttt{M;PL}} & {\small\texttt{F;SG}} & {\foreignlanguage{greek}{ειρηνική}} & pacifica, pacifico, \st{pacifici} \\
    {\foreignlanguage{greek}{ειρηνικό}} & {\small\texttt{Neut;NOM;SG}} & pacifica & {\small\texttt{F;SG}} & {\small\texttt{SG}} & {\foreignlanguage{greek}{ειρηνικό}} & pacifica, pacifico, \st{pacifici} \\
    {\foreignlanguage{greek}{ειρηνικά}} & {\small\texttt{Neut;NOM;PL}} &  & & {\small\texttt{PL}} &  {\foreignlanguage{greek}{ειρηνικά}} & pacifici, \st{pacifica}, \st{pacifico} \\
    \bottomrule
    \end{tabular}
    \caption{Triangulation and filtering example on Greek--Italian. All words are valid translations of the English word `peaceful'. We also show \st{filtered-out translations}.}
    \label{tab:triangulation}
    
\end{table*}


The typically used evaluation dictionaries cover a narrow breadth of the possible language pairs, with the majority of them focusing in pairs with English (as with the MUSE or \citet{dinu2015improving} dictionaries) or among high-resource European languages.
\citet{glavas-etal-2019-properly}, for instance, highlighted Anglocentricity as an issue, creating and evaluating on~28 dictionaries between~8 languages (Croatian, English, Finnish, French, German, Italian, Russian, Turkish) based on Google Translate. In addition, \citet{czarnowska2019don} focused on the morphology dimension, creating morphologically complete dictionaries for~2 sets of~5 \textit{genetically related} languages (Romance: French, Spanish, Italian, Portuguese, Catalan; and Slavic: Polish, Czech, Slovak, Russian, Ukrainian).
In contrast to these two (very valuable!) works, our method for creating dictionaries for low-resource languages (\S\ref{sec:lrld}) leverages resources that are available for about~300 languages. In addition, we propose a simple triangulation process (\S\ref{sec:triangle}), that makes it possible to create dictionaries for arbitrary language pairs, given that dictionaries into a pivot language (usually English) are available for both languages.



\subsection{Low-Resource Language Dictionaries}
\label{sec:lrld}

Our approach for constructing dictionaries is straightforward, inspired by phrase table extraction techniques from phrase-based MT \citep{koehn2009statistical}.
This is an automatic process, and introduces some degree of noise.
Rather than controlling this through manual inspection, which would be impossible for all language pairs, we rely on fairly simple heuristics for controlling the dictionaries' quality.

The first step is collecting publicly available parallel data between English and the low-resource language of interest.
We use data from the TED \citep{qi18naacl}, OpenSubtitles \citep{lison2016opensubtitles2015}, 
WikiMatrix \citep{schwenk2019wikimatrix}, bible \citep{malaviya17emnlp},
and JW300 \citep{agic2019jw300} datasets.\footnote{Not all languages are available in all these datasets.} This results in 354k, 53k, and 623k English-to-X parallel sentences for Azerbaijani (\az), Belarusian (\be), and Galician (\gl) respectively.\footnote{The anglocentricity in this step is by necessity -- it is hard to find a large volume of parallel data in a language pair excluding English.}
We align the parallel sentences using \texttt{fast\_align} \citep{dyer2013simple}, and extract symmetrized alignments using the \texttt{gdfa} heuristic \citep{koehn2005edinburgh}.
In order to ensure that we do not extract highly domain-specific word pairs, we only use the TED, OpenSubtitles, and WikiMatrix parts for word-pair extraction.
Also, in order to control for quality, we only extract word pairs if they appear in the dataset more than~5 times, \emph{and} if the symmetrized alignment probability is higher than~30\% in both directions. 

\begin{figure}
\begin{center}
\resizebox{.35\textwidth}{!}{
\begin{tikzpicture}
\begin{axis}[
    width={5cm}, 
	nodes near coords,
	nodes near coords align={vertical},
	point meta=explicit symbolic,
    axis line style={draw=none},
    ticks=none,
    tick style={draw=none},
]
\end{axis}
\draw (-2,4) node [above] {{\Large \pt:}};
\draw (-2,3) node [above] {{\Large \en:}};
\draw (-2,2) node [above] {{\Large \cs:}};
\draw [very thick] (2,4)--(0.75,3.6);
\draw [very thick] (2,4)--(3.25,3.6);
\draw [very thick] (0.75,3)--(-0.5,2.6);
\draw [very thick] (0.75,3)--(1.95,2.6);
\draw [very thick] (3.25,3)--(2.05,2.6);
\draw [very thick] (3.25,3)--(4.5,2.6);
\draw (2,4) node [above] {{\Large \texttt{trabalho}}};
\draw (0.75,3) node [above] {{\Large \texttt{job}}};
\draw (3.25,3) node [above] {{\Large \texttt{work}}};
\draw (-0.5,2) node [above] {{\Large \texttt{pr\'{a}cu}}};
\draw (-0.5,1.5) node [above] {{\Large \texttt{zamestnanie}}};
\draw (2,2) node [above] {{\Large \texttt{praca}}};
\draw (2,1.5) node [above] {{\Large \texttt{pr\'{a}ca}}};
\draw (4.5,2) node [above] {{\Large \texttt{dielo}}};
\draw (4.5,1.5) node [above] {{\Large \texttt{pr\'{a}ce}}};
\draw (4.5,1.0) node [above] {{\Large \texttt{pracovn\'{e}}}};
\end{tikzpicture}
}
    \vspace{-8mm}
    \caption{Transitivity example.}
    \label{fig:transitivity}
\end{center}
\end{figure}
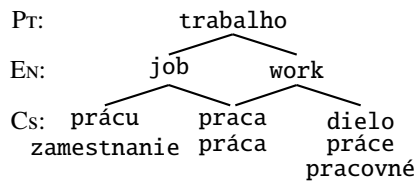
With this process, we end up with about 6k, 7k, and 38k word pairs for \az--\en, \be--\en, and \gl--\en{} respectively.
Following standard conventions, we sort the word pairs according to source-side frequency, and use the intermediate-frequency ones for evaluation, typically using the [5000--6500) rank boundaries.
The same process can be followed for any language pair with a sufficient volume of parallel data (needed for training a reasonably accurate word alignment model).\footnote{In fact, we can produce similar dictionaries for a large number of languages, as the combination of the recently created JW300 and WikiMatrix datasets provide an average of more than~100k parallel sentences in~300 languages. Before publication, we plan to create these dictionaries and make them publicly available, along with the corresponding code.}

\subsection{Dictionaries for all Language Pairs through Triangulation} 
\label{sec:triangle}
Our second method for creating new dictionaries is inspired by phrase table triangulation ideas from the pre-neural MT community \citep{wang-etal-2006-word,levinboim-chiang-2015-multi}.
The concept can be easily explained with an example, visualized in Figure~\ref{fig:transitivity}. 
Consider the Portuguese (\pt)  word \texttt{trabalho} which, according to the MUSE \pt--\en{} dictionary, has the words \texttt{job} and \texttt{work} as possible \en{} translations. In turn, these two \en{} words can be translated to~4 and~5 Czech (\cs) words respectively. 
By utilizing the \textit{transitive} property (which translation should exhibit) we can identify the set of~7 possible \cs{} translations for the \pt{} word \texttt{trabalho}.
Following this simple triangulation approach, we create~4,704 new dictionaries over pairs between the~50 languages of the MUSE dictionaries.\footnote{Available at \url{https://github.com/antonisa/embeddings}.} 
For consistency, we keep the same train and test splits as with MUSE, so that the source-side types are equal across all dictionaries with the same source language.

Triangulating through English (which is unavoidable, due to the relative paucity of non-English-centric dictionaries) is suboptimal -- English is morphologically poor and lacks corresponding markings for gender, case, or other features that are explicitly marked in many languages. As a result, several inflected forms in morphologically-rich languages map to the same English form.
Similarly, gendered nouns or adjectives in gendered languages map to English forms that lack gender information. For example, the MUSE Greek--English dictionary lists the word \texttt{peaceful} as the translation for all {\foreignlanguage{greek}{ειρηνικός, ειρηνική, ειρηνικό, ειρηνικά}}, which are the male, female, and neutral (singular and plural) inflections of the same adjective. Equivalently, the English--Italian dictionary translates \texttt{peaceful} into either \texttt{pacifico}, \texttt{pacifici}, or \texttt{pacifica} (male singular, male plural, and female singular, respectively; see Table~\ref{tab:triangulation}). 
When translating from or into English \textit{lacking context}, all of those are reasonable translations. 
When translating between Greek and Italian, though, one should at least take number into account (grammatical gender is a more complicated matter: it is not uncommon for word translations to be of different grammatical gender across languages). 

\begin{table*}[ht!]
    \centering
    {\small
    \setlength{\tabcolsep}{5pt}
    \begin{tabular}{@{}c@{}|c@{}cc@{}cc@{}cc@{}cc@{}cc@{}cc@{}cc@{}cc@{}cc@{}c|@{}c@{}c@{}}
    \toprule
\multirow{2}{*}{{\small src}} &  \multicolumn{20}{c}{{\small Target}}&&\\
& \multicolumn{2}{c}{\az} & \multicolumn{2}{c}{\be} & \multicolumn{2}{c}{\cs} & \multicolumn{2}{c}{\en} & \multicolumn{2}{c}{\es} & \multicolumn{2}{c}{\gl} & \multicolumn{2}{c}{\pt} & \multicolumn{2}{c}{\ru} & \multicolumn{2}{c}{\sk} & \multicolumn{2}{c}{\tr} & \ \ $\mu$\textsuperscript{\textsc{best}} \ \ \ & $\mu$\textsuperscript{\en}\\ \midrule
{\az}& -- & & 17.2 & \textsuperscript{\en} & 35.1 & \textsuperscript{\es} & 35.7 & \textsuperscript{\es} & 48.0 & \textsuperscript{\tr} & 32.7 & \textsuperscript{\ru} & 41.5 & \textsuperscript{\en} & 29.8 & \textsuperscript{\pt} & 31.7 & \textsuperscript{\cs} & 32.0 & \textsuperscript{\pt} & 33.7 & 31.7 \\
\be & 14.1 & \textsuperscript{\cs}& -- & & 35.9 & \textsuperscript{\tr} & \multicolumn{2}{c}{\cellcolor{gray!20}{29.9\textsuperscript{\pt}}} & \multicolumn{2}{c}{\cellcolor{gray!20}{39.5\textsuperscript{\en}}} & \multicolumn{2}{c}{\cellcolor{gray!20}{25.8\textsuperscript{\es}}} & \multicolumn{2}{c}{\cellcolor{gray!20}{34.4\textsuperscript{\es}}} & \multicolumn{2}{c}{\cellcolor{gray!20}{41.1\textsuperscript{\gl}}} & 30.7 & \textsuperscript{\ru} & 20.4 & \textsuperscript{\pt} & 30.2 & 28.8\\
\cs & 6.9 & \textsuperscript{\es} & 9.3 & \textsuperscript{\ru}& -- &  & 61.0 & \textsuperscript{\es} & 60.5 & \textsuperscript{\en} & \multicolumn{2}{c}{\cellcolor{gray!20}{27.9\textsuperscript{\pt}}} & 57.8 & \textsuperscript{\en} & 45.9 & \textsuperscript{\pt} & \multicolumn{2}{c}{\cellcolor{gray!85}{71.2\textsuperscript{\en}}} & 35.8 & \textsuperscript{\sk} & 41.8 & 41.2\\
\en & \multicolumn{2}{c}{\cellcolor{gray!20}{17.9\textsuperscript{\es}}} & 18.4 & \textsuperscript{\es} & 50.2 & \textsuperscript{\es}& -- &  & \multicolumn{2}{c}{\cellcolor{gray!85}{77.5\textsuperscript{\ru}}} & \multicolumn{2}{c}{\cellcolor{gray!20}{36.3\textsuperscript{\es}}} & \multicolumn{2}{c}{\cellcolor{gray!85}{72.3\textsuperscript{\sk}}} & 43.3 & \textsuperscript{\pt} & 40.4 & \textsuperscript{\tr} & 41.9 & \textsuperscript{\pt} & 44.2 & 42.7\\
\es & 12.1 & \textsuperscript{\en} & 10.1 & \textsuperscript{\ru} & 47.4 & \textsuperscript{\pt} & \multicolumn{2}{c}{\cellcolor{gray!85}{74.6\textsuperscript{\sk}}}& -- &  & \multicolumn{2}{c}{\cellcolor{gray!20}{37.5\textsuperscript{\es}}} & \multicolumn{2}{c}{\cellcolor{gray!85}{83.1\textsuperscript{\gl}}} & 41.9 & \textsuperscript{\tr} & 40.0 & \textsuperscript{\es} & 38.6 & \textsuperscript{\sk} & 42.8 & 41.4\\
\gl & 5.5 & \textsuperscript{\en} & 3.6 & \textsuperscript{\az} & \multicolumn{2}{c}{\cellcolor{gray!20}{26.5\textsuperscript{\tr}}} & \multicolumn{2}{c}{\cellcolor{gray!20}{43.2\textsuperscript{\es}}} & \multicolumn{2}{c}{\cellcolor{gray!85}{60.8\textsuperscript{\tr}}}& -- &  & \multicolumn{2}{c}{\cellcolor{gray!85}{52.9\textsuperscript{\cs}}} & \multicolumn{2}{c}{\cellcolor{gray!20}{23.8\textsuperscript{\tr}}} & \multicolumn{2}{c}{\cellcolor{gray!20}{26.8\textsuperscript{\cs}}} & \multicolumn{2}{c}{\cellcolor{gray!20}{19.7\textsuperscript{\cs}}} & 29.2 & 27.7\\
\pt & 5.8 & \textsuperscript{\pt} & 8.6 & \textsuperscript{\sk} & 47.2 & \textsuperscript{\gl} & \multicolumn{2}{c}{\cellcolor{gray!85}{71.3\textsuperscript{\en}}} & \multicolumn{2}{c}{\cellcolor{gray!85}{88.1\textsuperscript{\pt}}} & \multicolumn{2}{c}{\cellcolor{gray!20}{37.1\textsuperscript{\es}}}& -- &  & 38.0 & \textsuperscript{\es} & 38.7 & \textsuperscript{\es} & 38.1 & \textsuperscript{\en} & 41.4 & 40.4\\
\ru & 8.7 & \textsuperscript{\es} & 12.8 & \textsuperscript{\az} & 50.3 & \textsuperscript{\gl} & 55.5 & \textsuperscript{\tr} & 54.8 & \textsuperscript{\cs} & \multicolumn{2}{c}{\cellcolor{gray!20}{23.0\textsuperscript{\pt}}} & 52.4 & \textsuperscript{\en}& -- &  & \multicolumn{2}{c}{\cellcolor{gray!85}{45.5\textsuperscript{\tr}}} & 27.0 & \textsuperscript{\be} & 36.7 & 35.9\\
\sk & 4.0 & \textsuperscript{\be} & 10.9 & \textsuperscript{\ru} & \multicolumn{2}{c}{\cellcolor{gray!85}{72.5\textsuperscript{\be}}} & 55.6 & \textsuperscript{\tr} & 53.9 & \textsuperscript{\en} & \multicolumn{2}{c}{\cellcolor{gray!20}{28.4\textsuperscript{\en}}} & 52.0 & \textsuperscript{\es} & 44.0 & \textsuperscript{\gl}& -- &  & 28.5 & \textsuperscript{\en} & 38.9 & 37.9\\
\tr & 12.1 & \textsuperscript{\sk} & 9.0 & \textsuperscript{\az} & 41.8 & \textsuperscript{\ru} & 51.1 & \textsuperscript{\cs} & 55.0 & \textsuperscript{\en} & \multicolumn{2}{c}{\cellcolor{gray!20}{18.4\textsuperscript{\tr}}} & 51.6 & \textsuperscript{\en} & 34.6 & \textsuperscript{\en} & 29.4 & \textsuperscript{\es}& -- &  & 33.7 & 33.0\\

\midrule
$\mu$\textsuperscript{\textsc{best}} & \multicolumn{2}{c}{9.7} & \multicolumn{2}{c}{11.1} & \multicolumn{2}{c}{45.2} & \multicolumn{2}{c}{53.1} & \multicolumn{2}{c}{59.8} & \multicolumn{2}{c}{29.7} & \multicolumn{2}{c}{55.3} & \multicolumn{2}{c}{38.0} & \multicolumn{2}{c}{39.4} & \multicolumn{2}{c|}{31.3} & 37.3 \\
$\mu$\textsuperscript{\en} \ & \multicolumn{2}{c}{9.1} & \multicolumn{2}{c}{9.9} & \multicolumn{2}{c}{43.3} & \multicolumn{2}{c}{51.0} & \multicolumn{2}{c}{59.3} & \multicolumn{2}{c}{28.2} & \multicolumn{2}{c}{54.9} & \multicolumn{2}{c}{36.5} & \multicolumn{2}{c}{37.7} & \multicolumn{2}{c|}{30.8} & & 36.0\\
\bottomrule
\end{tabular}
}
    \caption{Lexicon Induction performance (measured with P@1) over~10 languages (90 pairs). In each cell, the superscript denotes the hub language that yields the best result for that language pair. $\mu$\textsuperscript{\textsc{best}}: average using the best hub language. $\mu$\textsuperscript{\en}: average using the \en{} as the hub. The lightly shaded cells are the language pairs where a bilingual \texttt{VecMap} system outperforms \texttt{MAT+MSPR}; in heavy shaded cells both \texttt{MUSEs} and \texttt{VecMap} outperform \texttt{MAT+MSPR}.}
    \label{tab:mwe}
\end{table*}

Hence, we devise a filtering method for removing blatant mistakes when triangulating morphologically rich languages.
We rely on automatic morphological tagging which we can obtain for most of the MUSE languages, using the StanfordNLP toolkit~\citep{qi2020stanza}.\footnote{The toolkit has since been renamed to Stanza. See \url{https://stanfordnlp.github.io/stanfordnlp/}.}
The morphological tagging uses the Universal Dependencies feature set~\citep{nivre2016universal} making the tagging comparable across almost all languages.
Our filtering technique iterates through the bridged dictionaries: for a given source word, if we find a target word with the exact same morphological analysis, we filter out all other translations with the same lemma but different tags. In the case of feature mismatch
(for instance, Greek uses~2 numbers,~4 cases and~3 genders while Italian has~2 numbers,~2 genders, and no cases)
or if we only find a partial tag match over a feature subset, we filter out translations with disagreeing tags.
We ignore the grammatical gender and verb form features, as they are not directly comparable cross-lingually.
Coming back to our Greek--Italian example, this means that for the form \foreignlanguage{greek}{ειρηνικός} we would only keep \texttt{pacifico} as a candidate translation (we show more examples in Table~\ref{tab:triangulation}).

Our filtering technique removes about~60.4\% of the entries in~2964 of the~4900 dictionaries.\footnote{Due to the lack of morphological analysis tools, we were unable to filter dictionaries in the following~11 languages: aze, bel, ben, bos, lit, mkd, msa, sqi, tam, tha, tel.} Unsurprisingly, we find that bridged dictionaries between morphologically rich languages require a lot more filtering. For instance more than~80\% of the entries of the Urdu-Greek dictionary get filtered out. On average, the languages with more filtered entries are Urdu (62.4\%), Turkish (61.1\%), and German (58.6\%). On the other hand, much fewer  entries are removed from dictionaries with languages like Dutch (36.2\%) or English (38.1\%).  Naturally, this filtering approach is restricted to languages for which a morphological analyzer is available. Mitigating this limitation is beyond the scope of this work, although it is unfortunately a common issue. For example, \citet{yova2019emnlp} manually corrected five dictionaries (between English and German, Danish, Bulgarian, Arabic, Hindi) but one needs to rely on automated annotations in order to scale to all languages.
Our method that uses automatically obtained morphological information combined with the guidelines proposed by \citet{yova2019emnlp} (e.g. removing proper nouns from the evaluation set) scales easily to multiple languages, allowing us to create more than~4 thousand dictionaries.

\section{Lexicon Induction Experiments}

The aim of our LI experiments is two-fold. First, the differences in LI performance show the importance of the hub language choice with respect to each evaluation pair. 
Second, as part of our call for moving beyond Anglo-centric evaluation, we also present LI results on several new language pairs using our triangulated dictionaries.


\subsection{Methods and Setup}

We train and evaluate all models starting with pre-trained Wikipedia FastText embeddings for all languages \citep{grave2018learning}. We focus on the minimally supervised scenario which only uses similar character strings between any languages for supervision in order to mirror the hard, realistic scenario of not having annotated training dictionaries between the languages. We learn MWE with the \texttt{MAT+MPSR} method 
using the publicly available code,\footnote{\url{https://github.com/ccsasuke/umwe}} aligning several language subsets varying the hub language.
We decided against comparing to the incremental hub (\texttt{IHS}) method of \citet{heyman-etal-2019-learning}, because the order in which the languages are added is an additional hyperparameter that would explode the experimental space.\footnote{We refer the reader to Table~2 from \citet{heyman-etal-2019-learning} which compares to \texttt{MAT+MPSR}, and to Table~7 of their appendix which shows the dramatic influence of language order.} We also do not compare to \texttt{UMH}, as we consider it conceptually similar to \texttt{MAT+MPSR} and no code is publicly available.
For BWE experiments, we use \texttt{MUSEs}\footnote{\url{https://github.com/facebookresearch/MUSE}} (MUSE, semisupervised)
and \texttt{VecMap}\footnote{\url{https://github.com/artetxem/vecmap}} systems, and we additionally compare them to \texttt{MAT+MPSR} for completeness.

We compare the statistical significance of the performance difference of two systems using paired bootstrap resampling \citep{koehn-2004-statistical}. Generally, a difference of 0.4--0.5 percentage points evaluated over our lexica is significant with $p<0.05$.

\paragraph{Experiment 1}
We first focus on~10 languages of varying morphological complexity and data availability (which affects the quality of the pre-trained word embeddings): Azerbaijani (\az), Belarusian (\be), Czech (\cs), English (\en), Galician (\gl), Portuguese (\pt), Russian (\ru), Slovak (\sk), Spanish (\es), and Turkish (\tr).
The choice of these languages additionally ensures that for our three low-resource languages (\az, \be, \gl) we include at least one related higher-resource language (\tr, \ru, \pt/\es{} respectively), allowing for comparative analysis.
Table~\ref{tab:mwe} summarizes the best post-hoc performing systems for this experiment.

\paragraph{Experiment 2}
In the second setting, we use a set of~7 more distant languages: English, French (\fr), Hindi (\hi), Korean (\ko), Russian, Swedish (\sv), and Ukrainian (\uk). This language subset has large variance in terms of typology and alphabet. The best performing systems are presented in Table~\ref{tab:mwe2}.

\subsection{Analysis and Takeaways}
\begin{table*}[t]
    \centering
    {\small
    \setlength{\tabcolsep}{5pt}
    \begin{tabular}{@{}c|c@{}cc@{}cc@{}cc@{}cc@{}cc@{}cc@{}c|cc@{}}
    \toprule
\multirow{2}{*}{{\small Source}} &  \multicolumn{14}{c}{{\small Target}}&&\\

& \multicolumn{2}{c}{\en} & \multicolumn{2}{c}{\fr} & \multicolumn{2}{c}{\hi} & \multicolumn{2}{c}{\ko} & \multicolumn{2}{c}{\ru} & \multicolumn{2}{c}{\sv} & \multicolumn{2}{c}{\uk} & $\mu$\textsuperscript{\textsc{best}} & $\mu$\textsuperscript{\en}\\ \midrule 
\en & -- &\textsuperscript{} & \multicolumn{2}{c}{\cellcolor{gray!20}{76.3\textsuperscript{\ru}}} & \multicolumn{2}{c}{\cellcolor{gray!85}{23.9\textsuperscript{\uk}}} & \multicolumn{2}{c}{\cellcolor{gray!85}{10.4\textsuperscript{\fr}}} & 42.0 & \textsuperscript{\uk} & \multicolumn{2}{c}{\cellcolor{gray!20}{59.0\textsuperscript{\hi}}} & 28.3 & \textsuperscript{\ru} & 40.0 & 38.5 \\
\fr & \multicolumn{2}{c}{\cellcolor{gray!20}{74.0\textsuperscript{\uk}}} & -- &\textsuperscript{} & \multicolumn{2}{c}{\cellcolor{gray!85}{19.0\textsuperscript{\ru}}}& \multicolumn{2}{c}{\cellcolor{gray!85}{7.5\textsuperscript{\sv}}} & 40.8 & \textsuperscript{\ru} & \multicolumn{2}{c}{\cellcolor{gray!20}{51.8\textsuperscript{\en}}} & 28.8 & \textsuperscript{\en} & 37.0 & 36.4\\
\hi & \multicolumn{2}{c}{\cellcolor{gray!85}{31.4\textsuperscript{\fr}}} & \multicolumn{2}{c}{\cellcolor{gray!85}{26.9\textsuperscript{\ru}}} & -- &\textsuperscript{} & \multicolumn{2}{c}{\cellcolor{gray!85}{2.1\textsuperscript{\en}}} & \multicolumn{2}{c}{\cellcolor{gray!85}{14.6\textsuperscript{\uk}}} & \multicolumn{2}{c}{\cellcolor{gray!85}{17.3\textsuperscript{\en}}} & \multicolumn{2}{c}{\cellcolor{gray!85}{10.5\textsuperscript{\fr}}} & 17.1 & 16.2\\
\ko & \multicolumn{2}{c}{\cellcolor{gray!85}{17.7\textsuperscript{\sv}}} & \multicolumn{2}{c}{\cellcolor{gray!85}{13.6\textsuperscript{\sv}}} & \multicolumn{2}{c}{\cellcolor{gray!85}{2.4\textsuperscript{\fr}}} & -- &\textsuperscript{} & \multicolumn{2}{c}{\cellcolor{gray!85}{7.9\textsuperscript{\en}}} & \multicolumn{2}{c}{\cellcolor{gray!85}{7.2\textsuperscript{\ru}}} & \multicolumn{2}{c}{\cellcolor{gray!85}{3.6\textsuperscript{\fr}}} & 8.8 & 7.9\\
\ru & \multicolumn{2}{c}{\cellcolor{gray!85}{53.4\textsuperscript{\ko}}} & \multicolumn{2}{c}{\cellcolor{gray!85}{51.7\textsuperscript{\ko}}} & \multicolumn{2}{c}{\cellcolor{gray!85}{15.3\textsuperscript{\uk}}} & \multicolumn{2}{c}{\cellcolor{gray!85}{5.2\textsuperscript{\en}}} & -- &\textsuperscript{} & \multicolumn{2}{c}{\cellcolor{gray!85}{41.3\textsuperscript{\uk}}} & \multicolumn{2}{c}{\cellcolor{gray!20}{56.3\textsuperscript{\ko}}} & 37.2 & 36.2\\
\sv & \multicolumn{2}{c}{\cellcolor{gray!20}{52.7\textsuperscript{\uk}}} & \multicolumn{2}{c}{\cellcolor{gray!20}{48.2\textsuperscript{\ko}}} & \multicolumn{2}{c}{\cellcolor{gray!85}{17.7\textsuperscript{\ru}}} & \multicolumn{2}{c}{\cellcolor{gray!85}{5.1\textsuperscript{\uk}}} & 33.2 & \textsuperscript{\fr} & -- &\textsuperscript{} & 24.1 & \textsuperscript{\ru} & 30.2 & 29.2 \\
\uk & \multicolumn{2}{c}{\cellcolor{gray!85}{41.4\textsuperscript{\ru}}} & \multicolumn{2}{c}{\cellcolor{gray!85}{44.0\textsuperscript{\hi}}} & \multicolumn{2}{c}{\cellcolor{gray!85}{14.4\textsuperscript{\sv}}} & \multicolumn{2}{c}{\cellcolor{gray!85}{2.6\textsuperscript{\en}}} & \multicolumn{2}{c}{\cellcolor{gray!20}{59.7\textsuperscript{\hi}}} & \multicolumn{2}{c}{\cellcolor{gray!85}{36.8\textsuperscript{\ko}}} & -- &\textsuperscript{} & 33.2 & 32.4 \\
\midrule
$\mu$\textsuperscript{\textsc{best}} & \multicolumn{2}{c}{45.1} & \multicolumn{2}{c}{43.5} & \multicolumn{2}{c}{15.5} & \multicolumn{2}{c}{5.5} & \multicolumn{2}{c}{33.0} & \multicolumn{2}{c}{35.6} & \multicolumn{2}{c|}{25.3} & 29.1 \\
$\mu$\textsuperscript{\en} & \multicolumn{2}{c}{42.7} & \multicolumn{2}{c}{42.5} & \multicolumn{2}{c}{14.5} & \multicolumn{2}{c}{5.1} & \multicolumn{2}{c}{32.4} & \multicolumn{2}{c}{34.9} & \multicolumn{2}{c|}{24.5} & & 28.1\\
\bottomrule
\end{tabular}
}
    \caption{Lexicon Induction performance (P@1) over MWEs from~7 typologically distant languages (42 pairs). The lightly shaded cells are the only language pairs where a bilingual \texttt{MUSE} system outperforms \texttt{MAT+MSPR}; in heavy shaded cells a bilingual \texttt{VecMap} (but not \texttt{MUSEs}) system outperform \texttt{MAT+MSPR}.}
    \label{tab:mwe2}
\end{table*}

\paragraph{MWE: English is rarely the best hub language}
In multilingual settings, we conclude that the standard practice of choosing English as the hub language is sub-optimal.
Out of the~90 evaluation pairs from our~10-language experiment (Table~\ref{tab:mwe}) the best hub language is English in only~17 instances (less than 20\% of the time). In fact, the average performance (over all evaluation pairs) when using \en{} as the hub (denoted as $\mu$\textsuperscript{\en}) is~1.3 percentage points worse than the optimal ($\mu$\textsuperscript{\textsc{best}}).
In our distant-languages experiment (Table~\ref{tab:mwe2}) English is the best choice only for~7 of the~42 evaluation pairs (again, less than 20\% of the time). As before, using \en{} as the hub leads to an average drop of one percentage point in performance aggregated over all pairs, compared to the averages of the optimal selection.
The rest of this section attempts to provide an explanation for these differences.

\paragraph{Expected gain for a hub language choice}

As vividly outlined by the superscript annotations in Tables~\ref{tab:mwe} and~\ref{tab:mwe2}, there is not a single hub language that stands out as the best one. Interestingly, all languages, across both experiments, are the best hub language for some evaluation language pair.
For example, in our~10-languages experiment, \es{} is the best choice for about~20\% of the evaluation pairs, \tr{} and \en{} are the best for about~17\% each, while \gl{} and \be{} are the best for only~5 and~3 language pairs respectively.

Clearly, not all languages are equally suited to be the hub language for many language pairs. Hence, it would be interesting to quantify how much better one could do by selecting the best hub language compared to a random choice. In order to achieve this, we define the expected gain $G_{l}$ of using language $l$ as follows.
Assume that we are interested in mapping $N$ languages into the shared space and $p^m_l$ is the accuracy\footnote{This could be substituted with any evaluation metric.} over a specified evaluation pair $m$ when using language $l$ as the hub. The random choice between $N$ languages will have an expected accuracy equal to the average accuracy when using all languages as hub: 
\begin{equation*}
\E [p^m] =\frac{\sum_l p^m_{l}}{N}.
\end{equation*} 

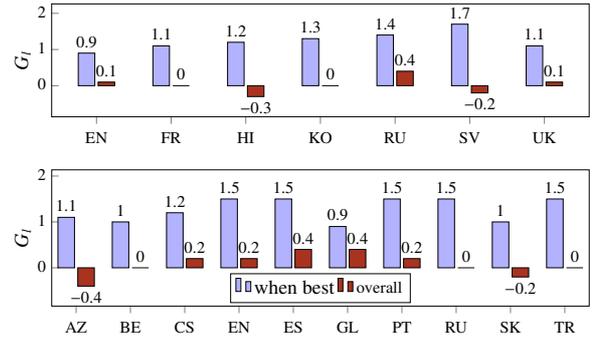
\begin{figure}
\begin{center}
    \resizebox{.5\textwidth}{!}{
\begin{tabular}{c}
\begin{tikzpicture}
\begin{axis}[
    height={4cm},
    width={13cm}, 
    ymin={-0.6},
    ymax={2},
	ylabel={{\large $G_l$}},
	ylabel shift={-0.1cm},
	enlargelimits=0.1,
	ybar=2pt,
	bar width=10pt,
	nodes near coords,
	nodes near coords align={vertical},
	point meta=y, 
	symbolic x coords={EN,FR,HI,KO,RU,SV,UK},
	xtick=data,
	typeset ticklabels with strut,
	xtick pos=left,
    ytick pos=left
]
\addplot [fill=blue!30!]
	coordinates {(EN,0.9) (FR,1.1) (HI,1.2) (KO,1.3) (RU,1.4) (SV,1.7) (UK,1.1)};
\addplot [fill=Mahogany]
	coordinates { (EN,0.1) (FR,0) (HI,-0.3) (KO,0) (RU,0.4) (SV,-0.2) (UK,0.1)};
\end{axis}
\end{tikzpicture}\\
\begin{tikzpicture}
\begin{axis}[
    height={4.5cm},
    width={13cm}, 
    ymin={-0.7},
    ymax={2},
	ylabel={{\large $G_l$}},
	ylabel shift={-0.1cm},
	enlargelimits=0.05,
	legend style={at={(0.5,.24)},
		anchor=north,legend columns=-1},
	ybar=2pt,
	bar width=10pt,
	nodes near coords,
	nodes near coords align={vertical},
	point meta=y, 
	symbolic x coords={AZ,BE,CS,EN,ES,GL,PT,RU,SK,TR},
	xtick=data,
	typeset ticklabels with strut,
	xtick pos=left,
    ytick pos=left
]
\addplot [fill=blue!30!]
	coordinates {(AZ,1.1) (BE,1.0) (CS,1.2) (EN,1.5) (ES,1.5) (GL,0.9) (PT,1.5) (RU,1.5) (SK,1.0) (TR,1.5)};
\addplot [fill=Mahogany]
	coordinates { (AZ,-0.4) (BE,0) (CS,0.2) (EN,0.2) (ES,0.4) (GL,0.4) (PT,0.2) (RU,0) (SK,-0.2) (TR,0)};
\legend{{\large when best,overall}}
\end{axis}
\end{tikzpicture}
\end{tabular}
}
    \caption{Expected gain $G_l$ for the MWE experiments.}
    \label{fig:gain}
\end{center}
\end{figure}

The gain for that evaluation dataset $m$ when using language $l$ as hub, then, is
$g_l^m = p^m_l - \E [p^m]$.
Now, for a collection of $M$ evaluation pairs we simply average their gains, in order to obtain the expected gain for using language $l$ as the hub:
\begin{equation*}
G_l = \E [g_l] = \frac{\sum_m g_l^m}{M}.
\end{equation*}

The results of this computation for both sets of experiments are presented in Figure~\ref{fig:gain}. The bars marked \texttt{`overall'} match our above definition, as they present the expected gain computed over all evaluation language pairs. For good measure, we also present the average gain per language aggregated over the evaluation pairs where that language was indeed the best hub language (\texttt{`when best'} bars).
Perhaps unsurprisingly, \az{} seems to be the worst hub language choice among the~10 languages of the first experiment, with an expected loss (negative gain) of~-0.4. This can be attributed to how distant \az{} is from all other languages, as well as to the fact that the \az{} pre-trained embeddings are of lower quality compared to all other languages (as the \az{} Wikipedia dataset is significantly smaller than the others).
Similarly, \hi{} and \sv{} show expected loss for our second experiment.

Note that English is not a \emph{bad} hub choice \textit{per se} -- it exhibits a positive expected gain in both sets of experiments. However, there are languages with larger expected gains, like \es{} and \gl{} in the~10-languages experiment that have a twice-as-large expected gain, while \ru{} has a 4 times larger expected gain in the distant-languages experiment.
Of course, the language subset composition of these experiments could possibly impact those numbers. For example, there are three very related languages (\es, \gl, \pt) in the~10 languages set, which might boost the expected gain for that subset; however, the trends stand even if we compute the expected gain over a subset of the evaluation pairs, removing all pairs that include \gl{} or \pt{}. For example, after removing all \gl{} results, \es{} has a slightly lower expected gain of $0.32$, but is still the language with the largest expected gain. 

\paragraph{Identifying the best hub language for a given evaluation set}
The next step is attempting to identify potential characteristics that will allow us make educated decisions with regards to choosing the hub language, given a specific evaluation set. For example, should one choose a language typologically similar to the evaluation source, target, or both? Or should they use the source or the target of the evaluation set as the hub?

Our first finding is that the best performing hub language will very likely be neither the source nor the target of the evaluation set. In our~10-languages experiments, a language different than the source and the target yields the best accuracy for over~93\% of the evaluation sets, with the difference being statistically significant in more than half such cases. Similarly, in the distant-languages experiment, there is only a single instance where the best performing hub language is either the source or the target evaluation language (for \fr--\ru), and for the other~97\% of cases the best option is a third language.
This surprising pattern contradicts the mathematical intuition discussed in Section~\ref{sec:mwe} according to which a model learning a single mapping (keeping another word embedding space fixed) is \textit{as expressive} as a model that learns two mappings for each of the languages. 
Instead, we find that in almost all cases, learning mappings for both language spaces of interest (hence rotating both spaces) leads to better BLI performance compared to when one of the spaces is fixed.

Our second finding is that the LI performance correlates with measures of distance between languages and language spaces.
The typological distance ($d^{gen}$) between two languages can be approximated through their genealogical distance over hypothesized language family trees, which we obtain from the URIEL typological database \citep{littell2017uriel}.
Also, \citet{patra19acl} recently motivated the use of Gromov-Hausdroff (GH) distance as an \textit{a priori} estimation of  how well two language embedding spaces can be aligned under an isometric transformation (an assumption most methods rely on).
The authors also note that vector space GH distance correlates with typological language distance.

\begin{figure}
\begin{center}
\resizebox{.35\textwidth}{!}{
\begin{tabular}{c}
\begin{tikzpicture}
\begin{axis}[
    height={6.5cm},
    width={12cm}, 
    ymin={39},
    ymax={44},
	ylabel={{\Large P@1}},
	ylabel shift={-0.1cm},
	xlabel={{\Large $\text{GH}^{hub}_\gl + \text{GH}^{hub}_\en$}},
	xlabel shift={-0.1cm},
	nodes near coords,
	nodes near coords align={vertical},
	point meta=explicit symbolic,
	xtick pos=left,
    ytick pos=left
]
\addplot+[only marks,fill=black]
	coordinates {
	(1.22,40.3) [AZ]
	(1.85,41.8) [BE]
	(1.65, 41.9) []
	(0.9, 39.6) [EN]
	(1.37, 43.2) [ES]
	(0.9,40.8) [GL]
	(1.15,41.5) [PT]
	(1.71,41.9) [RU]
	(1.61,41.6) [SK]
	(1.63,42.1) [TR]
	};
\addplot[blue,only marks,every node near coord/.append style={xshift=3pt,yshift=-15pt},]
	coordinates {
	(1.65, 41.9) [CS]
	};
\addplot[only marks] coordinates { (1.6,39.3) [{\Large $\rho=0.73$}]
(1.05,43) [{\Large Results on \gl--\en}]};
\end{axis}
\end{tikzpicture}\\

\begin{tikzpicture}
\begin{axis}[
    height={6.5cm},
    width={12cm}, 
    ymin={20.5},
    ymax={24.5},
	ylabel={{\Large P@1}},
	ylabel shift={-0.1cm},
	xlabel={{\Large $\text{GH}^{hub}_\en + \text{GH}^{hub}_\hi$}},
	xlabel shift={-0.1cm},
	nodes near coords,
	nodes near coords align={vertical},
	point meta=explicit symbolic,
	xtick pos=left,
    ytick pos=left
]
\addplot+[only marks,fill=black]
	coordinates {
	(0.98,20.9) []
	(1.7,23.5) [FR]
	(0.98, 21.0) [HI]
	(1.15, 21.4) [KO]
	(1.8, 23.5) [RU]
	(1.6, 21.4) [SV]
	(1.7,23.9) [UK]
	};
\addplot[blue,only marks,every node near coord/.append style={xshift=11pt,yshift=-7pt},]
	coordinates {
	(0.98, 20.9) [EN]
	};
\addplot[only marks] coordinates { (1.6,20.7) [{\Large $\rho=0.87$}]
(1.11,23.7) [{\Large Results on \en--\hi}]};
\end{axis}
\end{tikzpicture}

\end{tabular}
}
    \caption{The Lexicon Induction accuracy generally correlates positively with the GH distance of the source and target language vector spaces to the hub language.}
    \label{fig:corr}
\end{center}
\end{figure}
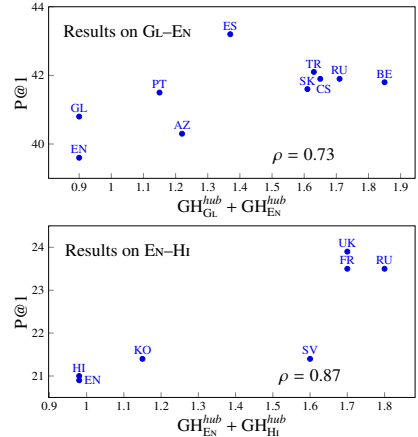
\begin{table*}[t]
    \centering
    \begin{tabular}[t]{l|r}
    \small
    \begin{tabular}{@{}rcccc|c@{}}
    \addlinespace[.1em]
    \toprule
        \multicolumn{4}{c}{Results on \az--\cs} & \multicolumn{2}{r}{Average}\\
    \midrule
    \multicolumn{2}{l}{\textit{Bilingual}}  & \az & \cs & & \multirow{2}{*}{25.8}\\
    \multicolumn{2}{l}{\ \ with hub:} & 22.7 & 29.1 & & \\
    \midrule
    \multicolumn{5}{l|}{\textit{Trilingual} \az, \cs, +hub:} & \\
    & \be & \en & \es & \gl & \multirow{4}{*}{28.2}\\
    & 21.6 & 28.5 & 31.8 & 23.0 \\
    & \pt & \ru & \sk & \tr \\
    & 29.6 & 27.4 & 30.4 & 32.9 \\
     \midrule
    \multicolumn{5}{l|}{\textit{Trilingual} \az, hub:\cs, +extra:}\\
    \multicolumn{1}{r}{\en{}} & \es &  \pt & \ru & \tr & \multirow{2}{*}{30.8}\\
    \multicolumn{1}{r}{30.1} & 30.1 & 33.2 & 27.1 & 33.7 \\
    \midrule
    \multicolumn{5}{l|}{\textit{Multilingual (10 languages)}} & \\
     \multicolumn{1}{r}{\az{}} & \be & \cs & \en & \es & \multirow{4}{*}{33.9}\\
     \multicolumn{1}{r}{33.7} & 34.0 & 32.3 & 34.5 & \textbf{35.1}\\
     \multicolumn{1}{r}{\gl{} } & \pt & \ru &\sk & \tr \\
     \multicolumn{1}{r}{34.0} & \textbf{34.8} & 34.5 & 32.9 & 33.7 \\
    \bottomrule
    \end{tabular}
    &
    \small
    \begin{tabular}{@{}rcccccc|c@{}}
    \toprule
        \multicolumn{6}{c}{Results on \ru--\uk} & \multicolumn{2}{r}{Average}\\
    \midrule
    \multicolumn{3}{l}{\textit{Bilingual}}  & \ru & \uk & & & \multirow{2}{*}{57.5}\\
    \multicolumn{3}{l}{\ \ with hub:} &  58.0 & 57.0 &  & & \\
    \addlinespace[0.1em]    
    \midrule
    \multicolumn{6}{l}{\textit{Trilingual} \be, \ru, \uk{}  with hub:} & & \\
    && \be & \ru & \uk  & & & \multirow{2}{*}{58.8}\\
    && \textbf{59.2} & \textbf{58.9} & 58.4 & & \\
    \midrule
    \multicolumn{6}{l}{\textit{Trilingual}  \ru, \uk{}, +hub:} & & \\
    \az & \cs & \en & \es & \fr & \hi & \tr & \multirow{2}{*}{57.8}\\
     57.4 & 58.5 & 58.4 & 58.3 & 58.0 & 57.0 & 57.2 & \\
    \addlinespace[0.2em]   
    \midrule
    \multicolumn{6}{l}{\textit{Multilingual \be, \ru, \uk, +hub:}} & \\
     \cs & \en & \es & \gl & \ko & \pt & \sv & \multirow{2}{*}{58.1}\\
     58.0 & 58.1 & 58.5 & \textbf{58.8} & 57.0 & 58.3 & 58.2 & \\
    \addlinespace[0.2em]
    \midrule
    \multicolumn{8}{l}{\textit{Multilingual \ru, \uk, \en, \fr, \hi, \ko, \sv, with hub:}} \\
     \en{} & \fr & \hi & \ko & \ru & \sv & \uk  & \multirow{2}{*}{55.6}\\
     55.3 & 56.1 & 55.8 & 56.3 & 55.3 & 55.3 & 54.9 & \\
    \bottomrule
    \end{tabular}\\
    \end{tabular}
    \caption{Comparison of bilingual, trilingual, and multilingual systems for distant (left) and related (right) languages. Multilinguality boosts performance significantly on distant languages.}
    \label{tab:focus}
\end{table*}
We find that there is a \textit{positive} correlation between LI performance and the genealogical distances between the source--hub and target--hub languages. The average (over all evaluation pairs) Pearson's correlation coefficient between P@1 and $d^{gen}$ is $0.49$ for the distant languages experiment and $0.38$ for the~10-languages one.
A similar positive correlation of performance and the sum of the GH distances between the source--hub and target--hub spaces. On our distant languages experiment, the correlation coefficient between P@1 and GH is 0.45, while it is slightly lower (0.34) for our~10-languages experiment. 
Figure~\ref{fig:corr} shows two high correlation examples, namely \gl--\en{} and \en--\hi{}.

\paragraph{BWE: The hub matters for distant languages}

\begin{table}
  \centering
    \begin{tabular}{@{}c|cc|c|cc@{}}
    \toprule
        \multirow{2}{*}{Test} & \multicolumn{2}{c|}{Hub} &         \multirow{2}{*}{Test} & \multicolumn{2}{c}{Hub}  \\
         & \textsc{src} & \textsc{trg} & & \textsc{src} & \textsc{trg}  \\
    \midrule 
        \az--\cs & 22.7 & 29.1 & \gl--\pt & 53.5 & 53.6 \\
        \az--\en & 13.2 & 20.7 & \pt--\gl & 39.0 & 36.7 \\
        \az--\tr & 30.1 & 30.1 & \uk--\ru & 61.6 & 61.8 \\
    \bottomrule
    \end{tabular}
    \caption{The hub is important for BWE between distant languages with 
    \texttt{MAT+MPSR}.}
    \label{tab:bil}
\end{table}

\texttt{MUSEs} implements a provably direction-independent closed form solution of the Procrustes problem, and we confirm  empirically that the hub choice does not affect the outcome (we provide complete results on \texttt{MUSEs} in Table~\ref{tab:app4} in the Appendix). 
Similarly, because \texttt{VecMap} uses symmetric re-weighting and produces bidirectional dictionaries at its final step, the results are not dependent on the training direction.
However, obtaining good performance with such methods requires the orthogonality assumption to hold, which for distant languages is rarely the case~\citep{patra19acl}. In fact, we find that the gradient-based \texttt{MAT+MPSR} method in a bilingual setting over typologically distant languages exhibits better performance than \texttt{MUSEs} or \texttt{VecMap}. Across Table~\ref{tab:mwe}, in only a handful of examples (shaded cells) do \texttt{VecMap} or \texttt{MUSEs} systems outperform \texttt{MAT+MPSR} for BWE (with the majority being among \en, \es, \gl, and \pt{}, all related high-resource languages). 

In the~7 distant languages setting, however, the results are different: \texttt{VecMap} outperforms \texttt{MUSEs} and the multilingual \texttt{MAT+MPSR} in the vast majority of the language pairs. The difference is more stark when the languages of the pair use completely different alphabets, where the same-character strings heuristic for bootstrapping the initial dictionary mapping fails. Instead, the monolingual similarity approach employed by \texttt{VecMap} is definitely more appropriate for settings such as those posed by languages like Korean or Hindi. 
This highlights the importance of actually evaluating and reporting results on such language pairs.

On the one hand, we find that when aligning distant languages with \texttt{MAT+MPSR}, the difference between hub choices can be significant -- in \az--\en{}, for instance, using \en{} as the hub leads to more than~7 percentage points difference compared to using \az{}. We show some examples in Table~\ref{tab:bil}.
On the other hand, when aligning typologically similar languages, the difference 
is less pronounced. For example, we obtain practically similar performance for \gl--\pt{}, \az--\tr{}, or \uk--\ru{} when using either the source or the target language as the hub. Note, though, that non-negligible differences could still occur, as in the case of \pt--\gl{}. 
In most cases, it is the case that the higher-resourced language is a better hub than the lower-resourced one, especially when the number of resources differ significantly (as in the case of \az{} and \be{} against any other language). Since BWE settings are not our main focus, we leave an extensive analysis of this observation for future work.

\paragraph{Bi-, tri-, and multilingual systems}
This part of our analysis compares bilingual, trilingual, and multilingual systems, with a focus on the under-represented languages. Through multiple experiments (complete evaluations are listed in the Appendix) we reach two main conclusions. On one hand, when evaluating on typologically distant languages, one should use as many languages as possible. In Table~\ref{tab:focus} we present one such example with results on \az--\cs{} under various settings.
On the other hand, when multiple related languages are available, one can achieve higher performance with multilingual systems containing all related languages and one more hub language, rather than learning diverse multilingual mappings using more languages. We confirm the latter observation with experiments on the Slavic (\be, \ru, \uk) and Iberian (\es, \gl, \pt) clusters, and present an example (\ru--\uk{}) in Table~\ref{tab:focus}.

\begin{table}[t]
    \centering
    \begin{tabular}{@{}cccc|ccc@{}}
    \toprule
    \multicolumn{4}{c|}{Transfer from \en} & \multicolumn{3}{c}{Transfer from \pt}\\
        Hub & \es{} & \pt{} & \gl{} & Hub & \es{} & \gl{} \\
    \midrule 
         \en & \textbf{38.7}& 21.8 & 19.4 & \en & \textbf{48.4} & 32.9 \\
        \es & 26.5& 16.1& \textbf{28.5}$^\dagger$ & \es & 41.4 & 25.5$^\dagger$ \\
        
        \pt & 28.1& 25.7& 15.6 & \pt & 44.3$^\dagger$ & \textbf{36.5} \\
        \gl & 35.4& 22.8& 23.1 & \gl & 48.1 & 23.8\\
        \be & 35.6& 30.5 & 13.2 & \\
        \ru & 28.6$^\dagger$ & \textbf{30.6} & 18.2 & \multicolumn{3}{l}{$^\dagger$: best train-test hub}\\
        \sk & 24.2& 30.2$^\dagger$ & 14.6 & \multicolumn{3}{l}{for LI.}\\
    \bottomrule
    \end{tabular}
    \caption{The choice of hub can significantly affect downstream zero-shot POS tagging accuracy.}
    \label{tab:pos}
\end{table}
\section{Downstream Task Experiments} Differences in BLI performance do not necessarily translate to differences in other downstream tasks that use the aligned embeddings, so ~\citet{glavas-etal-2019-properly} advocate for actual evaluation on such tasks.
We extend our analysis to an example downstream task of zero-shot POS tagging using the aligned embeddings for select language pairs.
We show that indeed the choice of the hub language can have dramatic impact.
Using Universal Dependencies data~\cite{nivre2016universal} we train simple bi-LSTM POS taggers on \en{} and \pt{} using the respective embeddings produced from each \texttt{MAT+MPSR} run, and evaluate the zero-shot performance on \gl{} and \es.\footnote{Note that our goal is not to achieve SOTA in zero-shot POS-tagging, but to show that embeddings resulting from different hub choices have different qualities.}
Although all taggers achieve consistent accuracies $>95\%$ on English and Portuguese regardless of the original \en{} or \pt{} embeddings, the zero-shot performance on the test languages, as shown in Table~\ref{tab:pos}, varies widely. 
For instance, using the embeddings produced from using \pt{} as a hub, we obtain the highest zero-shot accuracy on \gl{} (36.5\%), while using the ones from the \gl{} hub lead to significantly worse performance (23.8\%).
It should be noted that the best hub for POS-tagging does not always coincide with the best hub for LI, e.g. the best LI hub for \pt--\gl{} is \es{}, which leads to~11 percentage points worse \gl{} POS tagging performance than the best system. In fact, for the language pairs that we studied we observe no correlation between the two tasks performance as we vary the hub (with an average Spearman's rank correlation $\rho=0.08$).

\section{Conclusion}
With this work we challenge the standard practice in learning cross-lingual word embeddings. We empirically show that the choice of the hub language is an important parameter that affects lexicon induction performance in both bilingual (between distant languages) and multilingual settings.
More importantly, we hope that by providing new dictionaries and baseline results on several language pairs, we will stir the community towards evaluating all methods in challenging scenarios that include under-represented language pairs. Towards this end, our analysis provides insights and general directions for stronger baselines for non-Anglocentric cross-lingual word embeddings.
The problem of identifying the best \textit{hub} language, despite our analysis based on the use of typological distance, remains largely unsolved. In the future, we will investigate a hub language ranking/selection model a la \citet{lin-etal-2019-choosing}.

\section*{Acknowledgements}
The authors are grateful to the anonymous reviewers for their exceptionally constructive and insightful comments, and to Gabriela Weigel for her invaluable help with editing and proofreading the paper. This material is based upon work generously supported by the National Science Foundation under grant 1761548.

\bibliographystyle{acl_natbib}
\bibliography{References}

\clearpage
\newpage
\appendix

\section{Does evaluation directionality matter?}
We also explored whether there are significant differences between the evaluated quality of aligned spaces, when computed on both directions (src--trg and trg--src).
We find that the evaluation direction indeed matters a lot, when the languages of the evaluation pair are very distant, in terms of morphological complexity and data availability (which affects the quality of the original embeddings).
A prominent example, from our European-languages experiment, are evaluation pairs involving \az{} or \be{}.
When evaluating on the \az--XX and \be--XX dictionaries, the word translation P@1 is more than 20 percentage points higher than when evaluating on the opposite direction (XX-\az{} or XX-\be{}). For example, \es--\az{} has a mere P@1 of 9.9, while \az--\es{} achieves a P@1 of 44.9. This observation holds even between very related languages (cf. \ru--\be{}: 12.8, \be--\ru{}: 41.1 and \tr--\az: 8.4, \az--\tr: 32.0), which supports our hypothesis that this difference is also due to the quality of the pre-trained embeddings. It is important to note that such directionality differences are not observed  when evaluating distant pairs with presumably high-quality pre-trained embeddings e.g. \tr--\sk{} or \tr--\es{}; the P@1 for both directions is very close.

\section{Complete results for all experiments}
Here we provide complete evaluation results for our multilingual experiments. 
Table~\ref{tab:app4} presents the P@1 of the bilingual experiments using \texttt{MUSE}, and Table~\ref{tab:app5} presents accuracy using \texttt{VecMap}.
Tables~\ref{tab:app1}--\ref{tab:app2b}
present P@1, P@5, and P@10 respectively, for the experiment on the~10 European languages. Similarly, results on the distant languages experiment are shown in Tables~\ref{tab:app3a},~\ref{tab:app3b}, and~~\ref{tab:app3c}. 

\begin{table*}[t]
    \centering
    \caption{BWE results (P@1) with \texttt{MUSE}}
    \label{tab:app4}
    \setlength{\tabcolsep}{5pt}
\begin{tabular}{c}
\begin{tabular}{c|cccccccccc}
    \toprule
\multirow{2}{*}{{\small Source}} &  \multicolumn{10}{c}{{\small Target}}\\
& \multicolumn{1}{c}{\az} & \multicolumn{1}{c}{\be} & \multicolumn{1}{c}{\cs} & \multicolumn{1}{c}{\en} & \multicolumn{1}{c}{\es} & \multicolumn{1}{c}{\gl} & \multicolumn{1}{c}{\pt} & \multicolumn{1}{c}{\ru} & \multicolumn{1}{c}{\sk} & \multicolumn{1}{c}{\tr} \\ \midrule
\az & -- & 4.8 & 21.4 & 23.6 & 32.6 & 13.6 & 26.7 & 10.4 & 15.0 & 31.8\\
\be & 4.0 & -- & 26.1 & 3.8 & 12.3 & 9.3 & 11.3 & 42.0 & 23.1 & 2.9\\
\cs & 2.6 & 5.4 & -- & 57.1 & 55.5 & 11.9 & 52.3 & 44.7 & 71.2 & 31.6\\
\en & 12.2 & 2.5 & 47.3 & -- & 79.3 & 32.0 & 72.9 & 39.7 & 34.3 & 40.6\\
\es & 7.8 & 2.4 & 45.0 & 76.7 & -- & 37.1 & 83.4 & 38.9 & 34.3 & 38.2\\
\gl & 2.7 & 1.8 & 14.0 & 38.5 & 61.2 & -- & 53.3 & 11.4 & 12.9 & 8.5\\
\pt & 2.9 & 2.3 & 44.9 & 72.2 & 88.7 & 36.3 & -- & 33.7 & 33.7 & 34.6\\
\ru & 1.7 & 12.0 & 48.6 & 50.2 & 49.4 & 6.6 & 46.8 & -- & 44.6 & 21.1\\
\sk & 0.3 & 5.2 & 71.8 & 48.0 & 46.4 & 9.3 & 44.4 & 43.2 & -- & 21.2\\
\tr & 10.8 & 0.3 & 35.8 & 48.0 & 50.9 & 3.5 & 45.9 & 26.9 & 20.3 & --\\
\bottomrule
\end{tabular}
\\
\\[2em]
\begin{tabular}{c|ccccccc}
    \toprule
\multirow{2}{*}{{\small Source}} &  \multicolumn{7}{c}{{\small Target}}\\
& \multicolumn{1}{c}{\en} & \multicolumn{1}{c}{\fr} & \multicolumn{1}{c}{\hi} & \multicolumn{1}{c}{\ko} & \multicolumn{1}{c}{\ru} & \multicolumn{1}{c}{\sv} & \multicolumn{1}{c}{\uk} \\ \midrule
\en & -- & 80.3 & 17.9 & 9.5 & 39.7 & 60.0 & 25.9\\
\fr & 76.6 & -- & 11.9 & 5.1 & 38.0 & 52.4 & 26.8\\
\hi & 24.2 & 17.0 & -- & 0.4 & 3.1 & 3.3 & 2.3\\
\ko & 12.4 & 7.1 & 0.4 & -- & 2.5 & 2.2 & 0.6\\
\ru & 50.2 & 47.3 & 3.2 & 1.6 & -- & 35.8 & 58.8\\
\sv & 53.3 & 47.8 & 5.2 & 2.3 & 27.8 & -- & 19.9\\
\uk & 37.4 & 40.3 & 4.1 & 0.3 & 60.7 & 30.2 & --\\
\bottomrule
\end{tabular}
\end{tabular}

\end{table*}

\begin{table*}[t]
    \centering
    \caption{BWE results (P@1) with \texttt{VecMap}}
    \label{tab:app5}
    \setlength{\tabcolsep}{5pt}
\begin{tabular}{c}
\begin{tabular}{c|cccccccccc}
    \toprule
\multirow{2}{*}{{\small Source}} &  \multicolumn{10}{c}{{\small Target}}\\
& \multicolumn{1}{c}{\az} & \multicolumn{1}{c}{\be} & \multicolumn{1}{c}{\cs} & \multicolumn{1}{c}{\en} & \multicolumn{1}{c}{\es} & \multicolumn{1}{c}{\gl} & \multicolumn{1}{c}{\pt} & \multicolumn{1}{c}{\ru} & \multicolumn{1}{c}{\sk} & \multicolumn{1}{c}{\tr} \\ \midrule
\az & --  & 15.86 & 32.43 & 32.38 & 37.81 & 28.48 & 37.29 & 26.58 & 29.38 & 28.71 \\
\be & 14.41 & --  & 35.31 & 32.74 & 43.67 & 30.56 & 36.58 & 43.49 & 30.0 & 20.77 \\
\cs & 6.78 & 8.65 & --  & 57.45 & 56.75 & 35.66 & 54.09 & 44.59 & 73.49 & 34.75 \\
\en & 20.12 & 18.06 & 46.41 & --  & 69.91 & 40.83 & 63.5 & 40.13 & 40.19 & 37.7 \\
\es & 11.66 & 8.9 & 45.09 & 69.49 & --  & 39.37 & 81.19 & 40.52 & 40.7 & 39.89 \\
\gl & 5.34 & 2.44 & 29.14 & 46.11 & 58.44 & --  & 51.64 & 26.57 & 28.53 & 22.47 \\
\pt & 6.72 & 6.97 & 43.48 & 66.21 & 85.68 & 41.17 & --  & 38.29 & 39.81 & 36.61 \\
\ru & 8.06 & 10.33 & 52.43 & 59.03 & 59.29 & 29.87 & 55.55 & --  & 49.93 & 27.73 \\
\sk & 2.92 & 9.26 & 70.16 & 56.73 & 52.35 & 36.62 & 50.96 & 45.47 & --  & 31.23 \\
\tr & 14.2 & 9.74 & 42.37 & 45.51 & 50.21 & 28.06 & 49.11 & 32.33 & 34.57 & --  \\
\bottomrule
\end{tabular}
\\
\\[2em]
\begin{tabular}{c|ccccccc}
    \toprule
\multirow{2}{*}{{\small Source}} &  \multicolumn{7}{c}{{\small Target}}\\
& \multicolumn{1}{c}{\en} & \multicolumn{1}{c}{\fr} & \multicolumn{1}{c}{\hi} & \multicolumn{1}{c}{\ko} & \multicolumn{1}{c}{\ru} & \multicolumn{1}{c}{\sv} & \multicolumn{1}{c}{\uk} \\ \midrule
\en & --  & 69.82 & 35.0 & 19.2 & 40.56 & 56.49 & 23.63 \\
\fr & 68.44 & --  & 28.27 & 15.53 & 38.18 & 49.71 & 26.95 \\
\hi & 44.61 & 38.52 & --  & 14.01 & 20.39 & 26.26 & 14.72 \\
\ko & 32.69 & 18.32 & 12.93 & --  & 11.72 & 18.45 & 7.21 \\
\ru & 59.24 & 55.18 & 21.65 & 10.65 & --  & 47.58 & 55.12 \\
\sv & 51.94 & 46.92 & 27.46 & 12.66 & 34.29 & --  & 26.96 \\
\uk & 42.61 & 47.82 & 17.92 & 5.21 & 57.64 & 43.23 & --  \\
\bottomrule
\end{tabular}
\end{tabular}

\end{table*}

\begin{table*}[ht!]
    \centering
    \caption{All results from the European-languages MWE experiment: P@1 (part 1).}
    \label{tab:app1}
    \begin{tabular}{c|cccccccccc|c}
    \toprule
    \multirow{2}{*}{Test} & \multicolumn{10}{|c|}{Hub language} & \multirow{2}{*}{$\mu$} \\
 & \az & \be & \cs & \en & \es & \gl & \pt & \ru & \sk & \tr & \\ \midrule
\az--\be & 13.7 & 12.6 & 14.2 & 17.2 & 16.4 & 13.9 & 15.0 & 15.6 & 14.5 & 15.8 & 14.9 \\ 
\az--\cs & 33.7 & 34.0 & 32.3 & 34.5 & 35.1 & 34.0 & 34.8 & 34.5 & 32.9 & 33.7 & 33.9 \\ 
\az--\en & 31.1 & 34.7 & 32.8 & 32.6 & 35.7 & 34.2 & 33.6 & 33.6 & 34.0 & 33.2 & 33.5 \\ 
\az--\es & 42.7 & 46.6 & 45.2 & 46.1 & 44.9 & 44.4 & 44.9 & 43.3 & 46.1 & 48.0 & 45.2 \\ 
\az--\gl & 25.9 & 27.2 & 29.0 & 26.5 & 29.0 & 24.7 & 27.2 & 32.7 & 31.5 & 25.9 & 28.0 \\ 
\az--\pt & 37.5 & 41.5 & 39.3 & 41.5 & 39.8 & 39.0 & 39.8 & 41.5 & 38.5 & 40.0 & 39.8 \\ 
\az--\ru & 27.9 & 27.1 & 27.1 & 27.4 & 27.7 & 29.0 & 29.8 & 26.3 & 26.3 & 28.5 & 27.7 \\ 
\az--\sk & 28.8 & 30.1 & 31.7 & 29.1 & 30.4 & 30.4 & 28.8 & 28.5 & 29.5 & 30.4 & 29.8 \\ 
\az--\tr & 29.8 & 30.8 & 32.0 & 30.1 & 31.3 & 30.8 & 32.0 & 31.1 & 32.0 & 31.8 & 31.2 \\ 
\be--\az & 10.4 & 13.3 & 14.1 & 13.0 & 11.9 & 12.7 & 12.4 & 13.0 & 13.3 & 13.0 & 12.7 \\ 
\be--\cs & 30.5 & 31.6 & 33.3 & 33.0 & 30.8 & 31.6 & 32.5 & 32.2 & 33.0 & 35.9 & 32.5 \\ 
\be--\en & 24.8 & 26.5 & 27.8 & 27.8 & 28.2 & 24.8 & 29.9 & 28.2 & 26.5 & 25.6 & 27.0 \\ 
\be--\es & 36.4 & 38.1 & 36.4 & 39.5 & 35.5 & 38.1 & 39.0 & 37.0 & 36.1 & 34.4 & 37.0 \\ 
\be--\gl & 24.4 & 24.4 & 22.9 & 24.9 & 25.8 & 22.6 & 24.9 & 23.5 & 22.6 & 24.4 & 24.0 \\ 
\be--\pt & 33.2 & 33.2 & 32.7 & 33.7 & 34.4 & 31.7 & 33.9 & 31.7 & 31.9 & 31.4 & 32.8 \\ 
\be--\ru & 40.9 & 40.9 & 40.6 & 40.3 & 40.0 & 41.1 & 39.1 & 38.9 & 39.7 & 40.0 & 40.1 \\ 
\be--\sk & 30.1 & 27.7 & 30.7 & 27.4 & 28.6 & 29.2 & 28.9 & 30.7 & 27.7 & 27.4 & 28.8 \\ 
\be--\tr & 17.7 & 17.2 & 18.9 & 19.9 & 17.4 & 18.9 & 20.4 & 18.7 & 16.9 & 18.4 & 18.5 \\ 
\cs--\az & 3.5 & 4.6 & 4.9 & 6.0 & 6.9 & 4.9 & 3.7 & 4.9 & 4.0 & 6.0 & 4.9 \\ 
\cs--\be & 8.6 & 7.8 & 8.6 & 8.6 & 8.8 & 7.8 & 8.8 & 9.3 & 9.3 & 8.6 & 8.6 \\ 
\cs--\en & 59.7 & 60.5 & 59.4 & 59.2 & 61.0 & 60.4 & 60.1 & 59.7 & 60.2 & 58.8 & 59.9 \\ 
\cs--\es & 59.0 & 59.1 & 57.5 & 60.5 & 59.2 & 58.7 & 58.9 & 59.6 & 59.1 & 57.6 & 58.9 \\ 
\cs--\gl & 27.1 & 26.9 & 27.1 & 27.6 & 27.0 & 21.4 & 27.9 & 27.1 & 26.5 & 26.1 & 26.5 \\ 
\cs--\pt & 56.9 & 55.6 & 55.4 & 57.8 & 55.5 & 56.9 & 55.6 & 57.3 & 56.1 & 54.1 & 56.1 \\ 
\cs--\ru & 44.2 & 45.5 & 45.5 & 45.0 & 45.5 & 45.3 & 45.9 & 45.0 & 45.2 & 45.9 & 45.3 \\ 
\cs--\sk & 69.8 & 69.8 & 70.2 & 71.2 & 70.6 & 70.2 & 70.4 & 69.7 & 68.4 & 70.2 & 70.0 \\ 
\cs--\tr & 35.3 & 35.2 & 34.6 & 35.1 & 34.7 & 34.7 & 35.1 & 35.0 & 35.8 & 34.2 & 35.0 \\ 
\en--\az & 15.8 & 17.7 & 16.6 & 17.5 & 17.9 & 16.9 & 17.5 & 16.1 & 16.6 & 17.2 & 17.0 \\ 
\en--\be & 16.4 & 15.1 & 17.6 & 14.9 & 18.4 & 17.4 & 15.6 & 17.1 & 15.9 & 16.4 & 16.5 \\ 
\en--\cs & 49.2 & 49.0 & 47.6 & 47.4 & 50.2 & 49.8 & 50.1 & 48.3 & 48.8 & 49.3 & 49.0 \\ 
\en--\es & 76.3 & 77.5 & 77.2 & 77.0 & 76.8 & 76.5 & 76.6 & 77.5 & 77.3 & 76.6 & 76.9 \\ 
\en--\gl & 35.0 & 35.8 & 36.0 & 35.2 & 36.3 & 31.9 & 35.9 & 36.2 & 35.3 & 35.0 & 35.3 \\ 
\en--\pt & 71.3 & 71.8 & 71.3 & 72.1 & 71.5 & 72.0 & 71.0 & 71.5 & 72.3 & 71.3 & 71.6 \\ 
\en--\ru & 42.5 & 43.3 & 42.7 & 40.8 & 43.1 & 43.3 & 43.3 & 41.3 & 41.4 & 42.8 & 42.4 \\ 
\en--\sk & 38.7 & 39.6 & 40.2 & 38.0 & 40.4 & 39.3 & 38.5 & 38.6 & 36.8 & 40.4 & 39.0 \\ 
\en--\tr & 40.5 & 41.7 & 41.3 & 41.6 & 39.4 & 40.9 & 41.9 & 41.0 & 41.3 & 40.9 & 41.0 \\ 
\es--\az & 8.4 & 10.8 & 9.0 & 12.1 & 10.5 & 10.5 & 10.8 & 9.6 & 11.8 & 11.8 & 10.5 \\ 
\es--\be & 9.9 & 7.2 & 8.5 & 9.3 & 7.5 & 9.9 & 9.9 & 10.1 & 9.1 & 8.8 & 9.0 \\ 
\es--\cs & 45.3 & 46.0 & 44.2 & 43.4 & 45.8 & 45.5 & 47.4 & 46.3 & 45.4 & 44.7 & 45.4 \\ 
\es--\en & 73.0 & 74.5 & 73.8 & 73.2 & 74.0 & 74.1 & 73.1 & 73.5 & 74.6 & 73.6 & 73.7 \\ 
\es--\gl & 37.1 & 37.0 & 37.1 & 36.9 & 37.5 & 33.7 & 36.8 & 37.0 & 36.8 & 36.7 & 36.7 \\ 
\es--\pt & 82.1 & 82.9 & 82.7 & 83.0 & 83.1 & 83.1 & 82.5 & 83.0 & 82.9 & 83.0 & 82.8 \\ 
\es--\ru & 41.4 & 41.5 & 41.2 & 39.4 & 41.3 & 41.9 & 40.9 & 40.3 & 40.2 & 41.9 & 41.0 \\ 
\es--\sk & 37.0 & 39.2 & 38.8 & 37.4 & 40.0 & 39.2 & 39.5 & 39.5 & 35.2 & 38.8 & 38.5 \\ 
\es--\tr & 37.5 & 38.0 & 37.7 & 38.2 & 37.6 & 37.8 & 38.4 & 37.8 & 38.6 & 37.9 & 38.0 \\ 
    \bottomrule
    \end{tabular}
\end{table*}

\begin{table*}[t]
    \centering
    \caption{All results from the European-languages MWE experiment: P@1 (part 2).}
    \label{tab:app2}
    \begin{tabular}{c|cccccccccc|c}
    \toprule
    \multirow{2}{*}{Test} & \multicolumn{10}{|c|}{Hub language} & \multirow{2}{*}{$\mu$} \\
 & \az & \be & \cs & \en & \es & \gl & \pt & \ru & \sk & \tr & \\ \midrule
\gl--\az & 4.0 & 4.6 & 4.3 & 5.5 & 5.0 & 4.1 & 5.2 & 4.7 & 4.8 & 5.0 & 4.7 \\ 
\gl--\be & 3.6 & 3.0 & 2.4 & 3.0 & 3.0 & 2.4 & 3.0 & 2.4 & 1.2 & 3.0 & 2.7 \\ 
\gl--\cs & 23.2 & 25.7 & 25.0 & 23.8 & 26.5 & 23.0 & 25.6 & 25.4 & 25.6 & 26.5 & 25.0 \\ 
\gl--\en & 40.3 & 41.8 & 41.9 & 39.6 & 43.2 & 40.8 & 41.5 & 41.9 & 41.6 & 42.1 & 41.5 \\ 
\gl--\es & 60.0 & 60.5 & 60.1 & 59.9 & 60.4 & 59.0 & 60.0 & 60.3 & 59.6 & 60.8 & 60.1 \\ 
\gl--\pt & 52.5 & 52.5 & 52.9 & 52.0 & 52.0 & 50.4 & 52.5 & 51.9 & 52.1 & 52.0 & 52.1 \\ 
\gl--\ru & 22.5 & 22.7 & 22.9 & 21.7 & 23.3 & 21.9 & 23.7 & 22.7 & 22.5 & 23.8 & 22.8 \\ 
\gl--\sk & 26.0 & 26.3 & 26.8 & 25.6 & 26.4 & 23.4 & 25.5 & 25.1 & 23.2 & 26.4 & 25.5 \\ 
\gl--\tr & 18.5 & 19.3 & 19.7 & 18.6 & 17.8 & 18.3 & 18.9 & 19.2 & 19.4 & 17.6 & 18.7 \\ 
\pt--\az & 3.8 & 4.7 & 5.8 & 5.0 & 5.0 & 3.2 & 5.8 & 5.0 & 5.5 & 4.7 & 4.8 \\ 
\pt--\be & 7.3 & 5.3 & 7.3 & 7.3 & 6.1 & 7.1 & 6.8 & 6.1 & 8.6 & 7.1 & 6.9 \\ 
\pt--\cs & 45.5 & 47.0 & 46.3 & 45.0 & 45.5 & 47.2 & 45.5 & 46.7 & 46.5 & 45.6 & 46.1 \\ 
\pt--\en & 69.9 & 70.9 & 70.2 & 71.3 & 71.1 & 70.5 & 70.6 & 71.3 & 70.6 & 70.8 & 70.7 \\ 
\pt--\es & 87.4 & 88.1 & 87.7 & 87.6 & 88.0 & 87.4 & 88.1 & 87.8 & 87.6 & 88.1 & 87.8 \\ 
\pt--\gl & 35.7 & 36.9 & 36.3 & 36.3 & 37.1 & 32.7 & 36.0 & 35.9 & 35.2 & 36.4 & 35.8 \\ 
\pt--\ru & 37.4 & 37.7 & 36.4 & 36.5 & 38.0 & 38.0 & 36.2 & 37.0 & 37.1 & 37.4 & 37.2 \\ 
\pt--\sk & 37.6 & 37.0 & 37.3 & 36.7 & 38.7 & 37.7 & 38.3 & 37.9 & 33.6 & 38.0 & 37.3 \\ 
\pt--\tr & 36.5 & 37.4 & 37.2 & 38.1 & 35.9 & 36.4 & 35.5 & 37.2 & 36.2 & 36.3 & 36.7 \\ 
\ru--\az & 5.0 & 6.4 & 6.2 & 7.8 & 8.7 & 7.3 & 7.5 & 7.3 & 6.7 & 7.5 & 7.0 \\ 
\ru--\be & 12.8 & 9.9 & 10.7 & 11.5 & 11.2 & 11.0 & 11.5 & 12.3 & 11.0 & 11.8 & 11.4 \\ 
\ru--\cs & 49.2 & 50.0 & 49.2 & 50.1 & 49.7 & 50.3 & 50.3 & 49.8 & 50.1 & 50.1 & 49.9 \\ 
\ru--\en & 53.6 & 53.8 & 54.4 & 52.7 & 54.7 & 55.5 & 54.8 & 52.0 & 54.5 & 55.5 & 54.1 \\ 
\ru--\es & 53.7 & 53.4 & 54.8 & 54.5 & 52.3 & 53.5 & 54.0 & 53.2 & 53.9 & 51.2 & 53.4 \\ 
\ru--\gl & 20.9 & 21.3 & 22.1 & 22.3 & 22.9 & 17.2 & 23.0 & 21.8 & 21.7 & 21.9 & 21.5 \\ 
\ru--\pt & 50.4 & 50.3 & 50.4 & 52.4 & 51.1 & 51.1 & 49.6 & 49.8 & 51.0 & 47.6 & 50.4 \\ 
\ru--\sk & 45.0 & 44.7 & 44.7 & 45.2 & 45.2 & 44.7 & 44.3 & 43.7 & 43.7 & 45.5 & 44.7 \\ 
\ru--\tr & 25.9 & 27.0 & 26.2 & 26.9 & 26.0 & 25.9 & 26.1 & 25.6 & 26.8 & 24.7 & 26.1 \\ 
\sk--\az & 2.8 & 4.0 & 1.5 & 3.7 & 2.1 & 2.8 & 3.4 & 3.1 & 1.8 & 3.4 & 2.9 \\ 
\sk--\be & 10.2 & 7.5 & 9.9 & 9.4 & 9.6 & 8.3 & 10.4 & 10.9 & 10.9 & 9.1 & 9.6 \\ 
\sk--\cs & 71.4 & 72.5 & 70.9 & 70.8 & 70.5 & 71.1 & 71.3 & 70.6 & 71.0 & 71.4 & 71.1 \\ 
\sk--\en & 54.8 & 55.0 & 54.0 & 52.9 & 55.4 & 54.7 & 54.8 & 54.6 & 53.0 & 55.6 & 54.5 \\ 
\sk--\es & 52.5 & 51.6 & 52.2 & 53.9 & 52.3 & 52.0 & 50.4 & 50.5 & 51.5 & 51.1 & 51.8 \\ 
\sk--\gl & 27.0 & 27.3 & 27.2 & 28.4 & 27.8 & 20.6 & 26.2 & 26.0 & 27.0 & 27.0 & 26.4 \\ 
\sk--\pt & 49.3 & 50.3 & 48.2 & 50.4 & 52.0 & 49.2 & 49.1 & 48.7 & 48.5 & 47.7 & 49.3 \\ 
\sk--\ru & 43.8 & 43.4 & 43.5 & 43.2 & 43.7 & 44.0 & 42.8 & 42.9 & 41.2 & 43.4 & 43.2 \\ 
\sk--\tr & 28.2 & 27.5 & 27.2 & 28.5 & 27.1 & 26.1 & 26.2 & 27.6 & 27.4 & 26.0 & 27.2 \\ 
\tr--\az & 9.8 & 12.1 & 10.1 & 11.1 & 10.1 & 11.4 & 11.4 & 10.8 & 12.1 & 11.1 & 11.0 \\ 
\tr--\be & 9.0 & 4.8 & 8.7 & 8.1 & 7.8 & 7.5 & 8.1 & 6.9 & 7.5 & 7.2 & 7.6 \\ 
\tr--\cs & 40.3 & 41.6 & 40.3 & 41.6 & 41.6 & 40.8 & 41.6 & 41.8 & 40.9 & 39.2 & 41.0 \\ 
\tr--\en & 51.1 & 49.3 & 51.1 & 50.2 & 50.4 & 48.5 & 50.5 & 50.2 & 50.7 & 50.1 & 50.2 \\ 
\tr--\es & 53.8 & 53.6 & 55.0 & 55.0 & 52.5 & 53.0 & 54.6 & 52.9 & 54.1 & 53.3 & 53.8 \\ 
\tr--\gl & 17.0 & 17.3 & 17.3 & 15.9 & 16.8 & 11.6 & 17.5 & 17.1 & 17.1 & 18.4 & 16.6 \\ 
\tr--\pt & 50.1 & 50.1 & 51.4 & 51.6 & 49.3 & 48.9 & 48.7 & 49.9 & 50.5 & 49.5 & 50.0 \\ 
\tr--\ru & 34.0 & 34.3 & 32.3 & 34.6 & 34.3 & 33.6 & 33.2 & 32.0 & 33.0 & 32.9 & 33.4 \\ 
\tr--\sk & 27.5 & 29.2 & 27.9 & 28.5 & 29.4 & 27.7 & 27.9 & 27.5 & 25.2 & 27.9 & 27.9 \\ 
\bottomrule
    \end{tabular}
\end{table*}

\begin{table*}[t]
    \centering
    \caption{All results from the European-languages MWE experiment: P@5 (part 1).}
    \label{tab:app1a}
    \begin{tabular}{c|cccccccccc|c}
    \toprule
    \multirow{2}{*}{Test} & \multicolumn{10}{|c|}{Hub language} & \multirow{2}{*}{$\mu$} \\
 & \az & \be & \cs & \en & \es & \gl & \pt & \ru & \sk & \tr & \\ \midrule
 \az--\be & 26.0 & 22.5 & 26.5 & 26.0 & 26.5 & 25.2 & 25.7 & 26.0 & 25.7 & 25.7 & 25.6 \\ 
\az--\cs & 53.4 & 54.8 & 53.7 & 57.5 & 54.8 & 55.9 & 55.6 & 54.5 & 53.2 & 54.8 & 54.8 \\ 
\az--\en & 44.7 & 48.0 & 47.6 & 45.7 & 45.9 & 47.4 & 46.8 & 46.3 & 46.1 & 47.2 & 46.6 \\ 
\az--\es & 60.1 & 62.6 & 60.7 & 62.6 & 60.7 & 60.4 & 60.7 & 62.4 & 61.8 & 62.9 & 61.5 \\ 
\az--\gl & 38.3 & 37.7 & 40.1 & 41.4 & 38.9 & 35.8 & 40.1 & 41.4 & 38.9 & 39.5 & 39.2 \\ 
\az--\pt & 52.8 & 55.3 & 55.3 & 56.3 & 55.8 & 55.3 & 55.8 & 57.8 & 55.3 & 56.8 & 55.7 \\ 
\az--\ru & 45.2 & 46.5 & 46.8 & 48.1 & 49.2 & 47.3 & 48.4 & 45.5 & 46.8 & 50.0 & 47.4 \\ 
\az--\sk & 43.9 & 46.1 & 47.0 & 48.3 & 49.2 & 48.3 & 49.2 & 48.3 & 46.7 & 46.7 & 47.4 \\ 
\az--\tr & 45.2 & 49.1 & 51.3 & 49.1 & 46.7 & 48.7 & 49.1 & 49.4 & 49.6 & 49.4 & 48.8 \\ 
\be--\az & 20.6 & 20.6 & 23.4 & 23.2 & 24.6 & 22.0 & 22.9 & 24.9 & 22.3 & 24.6 & 22.9 \\ 
\be--\cs & 44.5 & 44.8 & 47.6 & 48.5 & 46.5 & 47.9 & 48.7 & 46.8 & 45.7 & 47.9 & 46.9 \\ 
\be--\en & 42.3 & 42.3 & 42.7 & 41.5 & 44.4 & 42.7 & 42.3 & 42.7 & 41.0 & 43.2 & 42.5 \\ 
\be--\es & 50.4 & 53.0 & 54.2 & 53.3 & 50.4 & 53.6 & 54.4 & 51.0 & 54.2 & 52.4 & 52.7 \\ 
\be--\gl & 38.8 & 36.5 & 37.7 & 38.8 & 38.0 & 36.5 & 38.3 & 38.0 & 38.6 & 37.7 & 37.9 \\ 
\be--\pt & 49.5 & 50.8 & 52.8 & 51.5 & 52.0 & 50.0 & 49.0 & 49.0 & 50.5 & 49.5 & 50.5 \\ 
\be--\ru & 53.0 & 53.2 & 52.1 & 51.8 & 53.8 & 52.7 & 53.0 & 53.0 & 53.2 & 51.8 & 52.8 \\ 
\be--\sk & 43.8 & 40.1 & 44.7 & 43.5 & 41.6 & 43.8 & 44.4 & 43.5 & 40.1 & 43.5 & 42.9 \\ 
\be--\tr & 33.4 & 33.2 & 34.6 & 37.8 & 32.2 & 34.4 & 36.9 & 33.4 & 33.2 & 32.2 & 34.1 \\ 
\cs--\az & 10.3 & 11.2 & 11.2 & 13.8 & 14.1 & 11.8 & 12.1 & 10.6 & 11.2 & 12.6 & 11.9 \\ 
\cs--\be & 14.8 & 15.5 & 15.5 & 16.3 & 16.3 & 16.6 & 16.1 & 16.1 & 14.8 & 15.8 & 15.8 \\ 
\cs--\en & 75.6 & 76.4 & 75.1 & 75.7 & 76.2 & 76.9 & 76.1 & 75.8 & 75.9 & 76.0 & 76.0 \\ 
\cs--\es & 75.5 & 75.3 & 74.1 & 76.5 & 75.9 & 74.9 & 74.3 & 75.5 & 75.9 & 74.1 & 75.2 \\ 
\cs--\gl & 40.8 & 41.8 & 43.0 & 43.7 & 43.1 & 36.5 & 42.1 & 42.6 & 42.1 & 41.2 & 41.7 \\ 
\cs--\pt & 72.9 & 74.1 & 72.2 & 74.3 & 73.1 & 73.7 & 72.7 & 73.8 & 72.7 & 71.6 & 73.1 \\ 
\cs--\ru & 64.5 & 64.4 & 63.6 & 63.9 & 63.9 & 64.5 & 64.9 & 64.5 & 64.3 & 65.5 & 64.4 \\ 
\cs--\sk & 81.7 & 82.9 & 83.2 & 82.8 & 82.5 & 83.0 & 83.2 & 82.7 & 81.6 & 82.7 & 82.6 \\ 
\cs--\tr & 56.2 & 56.0 & 55.1 & 57.1 & 56.4 & 54.2 & 54.9 & 55.5 & 54.9 & 53.8 & 55.4 \\ 
\en--\az & 28.3 & 29.1 & 30.3 & 29.9 & 28.9 & 29.2 & 30.2 & 29.1 & 28.8 & 30.6 & 29.4 \\ 
\en--\be & 32.8 & 28.3 & 34.0 & 31.5 & 34.0 & 34.5 & 30.3 & 32.8 & 33.3 & 32.8 & 32.4 \\ 
\en--\cs & 74.7 & 74.9 & 73.4 & 74.5 & 76.1 & 76.5 & 74.8 & 75.1 & 73.8 & 75.5 & 74.9 \\ 
\en--\es & 88.9 & 89.5 & 88.8 & 89.3 & 89.1 & 89.3 & 89.1 & 89.3 & 89.0 & 89.1 & 89.1 \\ 
\en--\gl & 49.0 & 50.4 & 50.5 & 50.4 & 51.3 & 47.8 & 50.9 & 51.4 & 49.1 & 50.7 & 50.1 \\ 
\en--\pt & 86.0 & 86.6 & 86.2 & 86.6 & 86.2 & 86.4 & 86.3 & 86.3 & 86.4 & 85.8 & 86.3 \\ 
\en--\ru & 68.0 & 68.1 & 68.2 & 66.0 & 68.6 & 69.6 & 68.7 & 67.7 & 67.4 & 68.2 & 68.1 \\ 
\en--\sk & 62.3 & 62.7 & 62.5 & 60.8 & 62.5 & 62.1 & 63.5 & 62.7 & 59.9 & 63.2 & 62.2 \\ 
\en--\tr & 63.6 & 62.6 & 64.3 & 62.4 & 62.4 & 63.8 & 63.8 & 63.0 & 63.2 & 63.2 & 63.2 \\ 
\es--\az & 16.3 & 16.9 & 16.9 & 17.5 & 18.4 & 17.8 & 17.2 & 17.2 & 19.0 & 18.1 & 17.5 \\ 
\es--\be & 16.8 & 15.5 & 17.1 & 18.9 & 16.3 & 18.9 & 18.7 & 17.1 & 18.1 & 16.5 & 17.4 \\ 
\es--\cs & 64.4 & 65.7 & 63.5 & 65.2 & 66.1 & 65.5 & 65.9 & 66.0 & 65.8 & 65.9 & 65.4 \\ 
\es--\en & 85.2 & 86.3 & 86.0 & 85.5 & 85.8 & 85.5 & 85.8 & 86.1 & 86.0 & 86.0 & 85.8 \\ 
\es--\gl & 45.6 & 46.0 & 45.7 & 46.1 & 46.4 & 43.2 & 45.9 & 45.7 & 45.8 & 46.2 & 45.7 \\ 
\es--\pt & 90.8 & 91.1 & 90.7 & 91.3 & 91.4 & 91.1 & 91.3 & 90.7 & 90.9 & 90.9 & 91.0 \\ 
\es--\ru & 61.5 & 62.5 & 61.4 & 62.5 & 62.1 & 61.7 & 62.2 & 60.8 & 61.6 & 62.9 & 61.9 \\ 
\es--\sk & 57.9 & 59.1 & 58.7 & 58.5 & 59.1 & 57.8 & 58.1 & 57.6 & 57.0 & 58.5 & 58.2 \\ 
\es--\tr & 57.0 & 57.4 & 57.2 & 56.7 & 55.0 & 56.3 & 56.3 & 55.5 & 56.6 & 56.5 & 56.5 \\ 
    \bottomrule
    \end{tabular}
\end{table*}

\begin{table*}[t]
    \centering
    \caption{All results from the European-languages MWE experiment: P@5 (part 2).}
    \label{tab:app2a}
    \begin{tabular}{c|cccccccccc|c}
    \toprule
    \multirow{2}{*}{Test} & \multicolumn{10}{|c|}{Hub language} & \multirow{2}{*}{$\mu$} \\
 & \az & \be & \cs & \en & \es & \gl & \pt & \ru & \sk & \tr & \\ \midrule
\gl--\az & 8.4 & 9.0 & 8.8 & 9.8 & 9.6 & 10.0 & 9.7 & 9.4 & 9.2 & 9.7 & 9.4 \\ 
\gl--\be & 7.3 & 6.1 & 6.1 & 6.7 & 6.7 & 6.7 & 7.9 & 6.1 & 6.1 & 7.3 & 6.7 \\ 
\gl--\cs & 41.8 & 42.1 & 43.0 & 42.3 & 44.5 & 40.2 & 42.5 & 42.5 & 42.0 & 43.0 & 42.4 \\ 
\gl--\en & 56.8 & 57.4 & 58.6 & 56.3 & 59.7 & 57.6 & 57.2 & 57.8 & 56.7 & 58.1 & 57.6 \\ 
\gl--\es & 68.3 & 68.8 & 68.1 & 68.8 & 68.6 & 67.9 & 68.3 & 68.8 & 68.2 & 68.8 & 68.5 \\ 
\gl--\pt & 63.9 & 64.3 & 63.4 & 64.1 & 63.2 & 62.8 & 63.4 & 64.0 & 63.7 & 63.9 & 63.7 \\ 
\gl--\ru & 40.2 & 39.8 & 39.3 & 39.6 & 39.5 & 37.0 & 40.0 & 39.5 & 39.3 & 40.8 & 39.5 \\ 
\gl--\sk & 41.6 & 42.4 & 41.1 & 41.9 & 43.7 & 38.5 & 41.0 & 41.4 & 39.2 & 41.5 & 41.2 \\ 
\gl--\tr & 33.5 & 33.4 & 34.9 & 33.9 & 33.3 & 29.4 & 32.4 & 32.6 & 34.0 & 31.5 & 32.9 \\ 
\pt--\az & 8.7 & 11.1 & 10.2 & 12.5 & 11.1 & 10.2 & 10.5 & 9.9 & 12.0 & 11.1 & 10.7 \\ 
\pt--\be & 14.4 & 12.1 & 14.4 & 17.4 & 14.1 & 15.9 & 14.9 & 14.9 & 14.9 & 14.6 & 14.8 \\ 
\pt--\cs & 65.6 & 66.6 & 64.7 & 65.8 & 66.5 & 66.6 & 65.9 & 66.3 & 65.5 & 65.1 & 65.9 \\ 
\pt--\en & 81.3 & 82.1 & 82.0 & 82.1 & 81.9 & 82.0 & 81.5 & 81.7 & 81.5 & 82.0 & 81.8 \\ 
\pt--\es & 92.1 & 92.6 & 92.4 & 92.1 & 92.0 & 91.8 & 92.4 & 92.4 & 92.0 & 92.3 & 92.2 \\ 
\pt--\gl & 45.4 & 46.4 & 46.2 & 46.9 & 46.8 & 43.5 & 45.8 & 45.4 & 45.2 & 46.7 & 45.8 \\ 
\pt--\ru & 57.6 & 57.8 & 57.7 & 58.7 & 58.1 & 58.5 & 57.0 & 57.5 & 57.6 & 57.6 & 57.8 \\ 
\pt--\sk & 57.2 & 56.9 & 57.0 & 57.8 & 56.6 & 55.4 & 56.6 & 56.8 & 53.1 & 56.4 & 56.4 \\ 
\pt--\tr & 53.9 & 54.8 & 54.2 & 56.3 & 53.3 & 53.6 & 52.7 & 54.5 & 54.4 & 54.6 & 54.2 \\ 
\ru--\az & 12.0 & 15.6 & 15.9 & 15.6 & 15.9 & 14.8 & 15.4 & 14.2 & 14.2 & 15.9 & 15.0 \\ 
\ru--\be & 20.1 & 18.3 & 20.6 & 20.1 & 20.9 & 20.6 & 20.6 & 20.9 & 21.1 & 20.4 & 20.4 \\ 
\ru--\cs & 65.7 & 65.0 & 65.1 & 64.7 & 65.0 & 66.7 & 66.1 & 65.8 & 65.1 & 65.5 & 65.5 \\ 
\ru--\en & 72.8 & 73.0 & 73.9 & 72.0 & 73.8 & 73.5 & 72.7 & 72.3 & 72.9 & 73.5 & 73.0 \\ 
\ru--\es & 70.1 & 69.8 & 69.7 & 71.3 & 69.2 & 70.3 & 71.2 & 68.8 & 70.7 & 68.4 & 69.9 \\ 
\ru--\gl & 36.1 & 35.9 & 36.1 & 36.8 & 37.1 & 30.9 & 36.5 & 36.6 & 35.9 & 35.3 & 35.7 \\ 
\ru--\pt & 66.8 & 66.8 & 67.0 & 69.3 & 67.9 & 67.6 & 65.8 & 66.6 & 67.3 & 65.2 & 67.0 \\ 
\ru--\sk & 61.1 & 62.6 & 61.4 & 61.1 & 62.0 & 61.8 & 61.8 & 60.9 & 59.8 & 61.6 & 61.4 \\ 
\ru--\tr & 48.0 & 48.0 & 47.6 & 49.9 & 47.1 & 47.5 & 48.0 & 46.0 & 47.0 & 47.4 & 47.7 \\ 
\sk--\az & 7.7 & 9.2 & 7.1 & 9.5 & 7.4 & 8.3 & 8.9 & 8.9 & 8.3 & 8.6 & 8.4 \\ 
\sk--\be & 17.4 & 16.7 & 18.5 & 18.2 & 17.7 & 18.5 & 18.2 & 19.3 & 19.3 & 18.5 & 18.2 \\ 
\sk--\cs & 82.1 & 82.1 & 81.3 & 81.6 & 82.1 & 82.4 & 81.6 & 81.6 & 81.3 & 81.9 & 81.8 \\ 
\sk--\en & 70.7 & 71.7 & 71.3 & 69.6 & 71.2 & 71.4 & 71.5 & 70.9 & 70.3 & 71.4 & 71.0 \\ 
\sk--\es & 69.2 & 69.7 & 70.2 & 71.2 & 70.1 & 68.8 & 70.0 & 68.6 & 69.2 & 69.4 & 69.6 \\ 
\sk--\gl & 43.4 & 43.3 & 42.9 & 45.1 & 43.7 & 36.0 & 42.9 & 42.0 & 43.0 & 42.7 & 42.5 \\ 
\sk--\pt & 68.2 & 67.5 & 67.5 & 68.7 & 69.9 & 67.6 & 66.1 & 67.6 & 66.7 & 66.7 & 67.7 \\ 
\sk--\ru & 59.2 & 58.1 & 58.2 & 58.8 & 59.4 & 59.5 & 58.8 & 58.5 & 57.5 & 59.5 & 58.8 \\ 
\sk--\tr & 47.2 & 48.7 & 47.6 & 48.7 & 47.1 & 46.7 & 48.2 & 47.8 & 46.7 & 46.2 & 47.5 \\ 
\tr--\az & 19.5 & 22.2 & 19.9 & 21.2 & 20.9 & 20.9 & 20.5 & 19.5 & 21.9 & 20.2 & 20.7 \\ 
\tr--\be & 17.1 & 12.3 & 16.2 & 17.1 & 16.8 & 15.6 & 16.5 & 16.5 & 16.2 & 16.2 & 16.1 \\ 
\tr--\cs & 61.6 & 62.1 & 60.1 & 61.8 & 62.4 & 61.9 & 61.6 & 61.5 & 61.4 & 60.1 & 61.4 \\ 
\tr--\en & 68.0 & 68.2 & 68.1 & 67.2 & 67.8 & 67.5 & 69.6 & 67.7 & 67.9 & 67.2 & 67.9 \\ 
\tr--\es & 69.8 & 69.0 & 70.4 & 70.5 & 68.0 & 69.2 & 70.5 & 69.4 & 69.8 & 69.5 & 69.6 \\ 
\tr--\gl & 30.5 & 30.7 & 31.1 & 30.0 & 30.4 & 23.6 & 31.4 & 31.1 & 29.7 & 30.7 & 29.9 \\ 
\tr--\pt & 67.1 & 66.9 & 66.9 & 67.9 & 66.5 & 65.9 & 65.2 & 67.1 & 67.5 & 66.6 & 66.8 \\ 
\tr--\ru & 55.4 & 55.9 & 54.0 & 55.4 & 55.3 & 55.1 & 55.1 & 53.0 & 52.9 & 53.5 & 54.6 \\ 
\tr--\sk & 48.2 & 49.9 & 48.9 & 49.7 & 48.7 & 47.8 & 48.9 & 48.1 & 44.2 & 47.7 & 48.2 \\  
\bottomrule
    \end{tabular}
\end{table*}

\begin{table*}[t]
    \centering
    \caption{All results from the European-languages MWE experiment: P@10 (part 1).}
    \label{tab:app1b}
    \begin{tabular}{c|cccccccccc|c}
    \toprule
    \multirow{2}{*}{Test} & \multicolumn{10}{|c|}{Hub language} & \multirow{2}{*}{$\mu$} \\
 & \az & \be & \cs & \en & \es & \gl & \pt & \ru & \sk & \tr & \\ \midrule
 \az--\be & 31.1 & 27.1 & 30.8 & 31.4 & 31.9 & 31.1 & 29.8 & 30.3 & 32.2 & 31.1 & 30.7 \\ 
\az--\cs & 60.3 & 62.5 & 60.8 & 62.7 & 63.6 & 61.4 & 62.7 & 61.1 & 60.3 & 63.6 & 61.9 \\ 
\az--\en & 49.3 & 51.1 & 52.6 & 50.5 & 49.5 & 50.7 & 51.4 & 50.3 & 50.1 & 50.7 & 50.6 \\ 
\az--\es & 63.8 & 65.7 & 65.4 & 67.1 & 65.2 & 66.3 & 68.0 & 64.6 & 66.6 & 67.4 & 66.0 \\ 
\az--\gl & 42.6 & 42.6 & 45.1 & 45.1 & 43.8 & 39.5 & 45.1 & 43.8 & 42.6 & 43.8 & 43.4 \\ 
\az--\pt & 58.5 & 61.2 & 62.7 & 62.5 & 61.5 & 61.7 & 61.0 & 61.2 & 61.7 & 62.5 & 61.5 \\ 
\az--\ru & 50.8 & 52.7 & 52.9 & 50.8 & 54.0 & 53.2 & 54.3 & 51.6 & 51.9 & 54.5 & 52.7 \\ 
\az--\sk & 48.9 & 52.0 & 53.0 & 52.0 & 53.9 & 54.2 & 53.0 & 52.4 & 51.7 & 51.7 & 52.3 \\ 
\az--\tr & 53.3 & 55.5 & 56.7 & 57.0 & 55.0 & 55.3 & 55.7 & 56.5 & 57.0 & 56.7 & 55.9 \\ 
\be--\az & 25.7 & 25.4 & 29.7 & 28.5 & 29.4 & 26.8 & 27.7 & 28.2 & 26.8 & 28.0 & 27.6 \\ 
\be--\cs & 50.7 & 51.0 & 52.1 & 51.3 & 51.8 & 53.8 & 52.7 & 51.8 & 50.7 & 51.8 & 51.8 \\ 
\be--\en & 46.6 & 48.7 & 50.0 & 46.2 & 48.3 & 50.9 & 46.2 & 48.3 & 46.2 & 47.9 & 47.9 \\ 
\be--\es & 54.7 & 57.3 & 58.7 & 58.7 & 56.2 & 57.9 & 57.9 & 55.9 & 58.5 & 57.9 & 57.4 \\ 
\be--\gl & 47.0 & 45.2 & 44.6 & 46.1 & 43.8 & 41.4 & 43.5 & 43.8 & 44.3 & 42.9 & 44.3 \\ 
\be--\pt & 55.3 & 55.8 & 57.0 & 57.8 & 57.0 & 56.5 & 55.8 & 54.5 & 55.5 & 56.0 & 56.1 \\ 
\be--\ru & 56.3 & 56.3 & 56.1 & 56.1 & 56.9 & 56.1 & 56.3 & 56.3 & 56.9 & 55.5 & 56.3 \\ 
\be--\sk & 48.0 & 45.6 & 48.3 & 47.7 & 48.0 & 48.6 & 49.8 & 48.6 & 46.2 & 48.0 & 47.9 \\ 
\be--\tr & 38.3 & 40.5 & 41.5 & 43.2 & 40.3 & 40.3 & 41.8 & 41.5 & 40.3 & 38.3 & 40.6 \\ 
\cs--\az & 13.8 & 14.9 & 15.5 & 16.1 & 17.5 & 14.9 & 15.8 & 14.1 & 14.9 & 15.5 & 15.3 \\ 
\cs--\be & 18.9 & 17.9 & 19.2 & 19.9 & 19.4 & 19.9 & 19.9 & 19.2 & 17.9 & 19.2 & 19.1 \\ 
\cs--\en & 80.2 & 80.5 & 79.8 & 80.0 & 80.1 & 81.0 & 80.2 & 80.5 & 80.5 & 81.1 & 80.4 \\ 
\cs--\es & 80.1 & 79.6 & 78.8 & 80.0 & 79.9 & 79.4 & 79.9 & 79.3 & 80.2 & 79.0 & 79.6 \\ 
\cs--\gl & 47.2 & 48.0 & 47.9 & 49.9 & 49.3 & 42.4 & 48.2 & 48.3 & 49.1 & 47.1 & 47.7 \\ 
\cs--\pt & 77.5 & 78.7 & 77.5 & 78.3 & 77.1 & 77.7 & 76.9 & 77.7 & 76.9 & 76.8 & 77.5 \\ 
\cs--\ru & 70.1 & 70.3 & 69.1 & 69.6 & 69.4 & 70.7 & 69.6 & 69.5 & 69.5 & 70.5 & 69.8 \\ 
\cs--\sk & 85.5 & 85.6 & 85.7 & 85.2 & 84.9 & 85.1 & 86.2 & 85.2 & 84.9 & 85.6 & 85.4 \\ 
\cs--\tr & 63.2 & 62.7 & 62.5 & 63.5 & 62.7 & 62.5 & 62.7 & 63.4 & 62.6 & 61.6 & 62.7 \\ 
\en--\az & 32.2 & 33.3 & 34.3 & 34.3 & 33.8 & 32.5 & 34.4 & 33.0 & 34.3 & 33.8 & 33.6 \\ 
\en--\be & 38.5 & 34.0 & 40.4 & 39.0 & 40.0 & 41.2 & 38.7 & 38.2 & 38.7 & 38.5 & 38.7 \\ 
\en--\cs & 81.2 & 81.1 & 79.9 & 80.7 & 81.9 & 82.5 & 80.6 & 80.7 & 80.7 & 81.5 & 81.1 \\ 
\en--\es & 91.3 & 92.1 & 91.7 & 91.5 & 91.9 & 91.7 & 91.8 & 91.6 & 91.9 & 91.7 & 91.7 \\ 
\en--\gl & 53.9 & 56.3 & 56.4 & 55.7 & 55.8 & 53.2 & 55.9 & 56.2 & 54.9 & 55.5 & 55.4 \\ 
\en--\pt & 89.4 & 90.0 & 89.2 & 89.5 & 89.1 & 89.5 & 89.3 & 89.0 & 89.4 & 89.0 & 89.3 \\ 
\en--\ru & 74.6 & 74.0 & 75.8 & 72.2 & 74.8 & 76.0 & 74.8 & 73.8 & 74.0 & 74.4 & 74.4 \\ 
\en--\sk & 69.3 & 69.7 & 69.9 & 68.0 & 69.6 & 68.7 & 69.9 & 69.5 & 67.1 & 69.9 & 69.2 \\ 
\en--\tr & 69.9 & 70.1 & 71.0 & 69.3 & 69.5 & 69.8 & 70.3 & 71.1 & 70.0 & 69.2 & 70.0 \\ 
\es--\az & 20.2 & 20.8 & 20.2 & 21.1 & 20.8 & 20.2 & 19.3 & 20.2 & 21.1 & 21.1 & 20.5 \\ 
\es--\be & 20.8 & 18.9 & 20.8 & 22.9 & 21.3 & 22.4 & 21.1 & 23.2 & 21.3 & 21.3 & 21.4 \\ 
\es--\cs & 70.5 & 70.7 & 70.8 & 70.9 & 71.0 & 71.1 & 71.3 & 71.8 & 72.2 & 70.9 & 71.1 \\ 
\es--\en & 88.5 & 88.4 & 88.5 & 88.3 & 88.5 & 88.5 & 88.5 & 88.5 & 88.5 & 88.4 & 88.5 \\ 
\es--\gl & 49.5 & 49.4 & 49.4 & 49.8 & 50.0 & 46.0 & 49.6 & 49.6 & 49.4 & 50.2 & 49.3 \\ 
\es--\pt & 92.7 & 92.5 & 92.5 & 92.5 & 93.0 & 92.9 & 92.8 & 92.4 & 92.1 & 92.7 & 92.6 \\ 
\es--\ru & 67.5 & 67.1 & 67.4 & 68.9 & 67.4 & 67.6 & 67.8 & 66.8 & 68.7 & 68.5 & 67.8 \\ 
\es--\sk & 64.5 & 64.3 & 63.9 & 65.4 & 65.4 & 63.5 & 64.3 & 64.8 & 63.0 & 63.8 & 64.3 \\ 
\es--\tr & 63.6 & 63.8 & 64.3 & 62.7 & 61.6 & 62.6 & 63.7 & 62.2 & 63.8 & 61.7 & 63.0 \\ 
    \bottomrule
    \end{tabular}
\end{table*}

\begin{table*}[t]
    \centering
    \caption{All results from the European-languages MWE experiment: P@10 (part 2).}
    \label{tab:app2b}
    \begin{tabular}{c|cccccccccc|c}
    \toprule
    \multirow{2}{*}{Test} & \multicolumn{10}{|c|}{Hub language} & \multirow{2}{*}{$\mu$} \\
 & \az & \be & \cs & \en & \es & \gl & \pt & \ru & \sk & \tr & \\ \midrule
 \gl--\az & 11.5 & 11.2 & 11.1 & 12.5 & 12.6 & 12.3 & 13.1 & 12.1 & 12.5 & 12.3 & 12.1 \\ 
\gl--\be & 8.5 & 7.3 & 8.5 & 9.1 & 8.5 & 7.9 & 7.9 & 7.9 & 8.5 & 9.7 & 8.4 \\ 
\gl--\cs & 48.0 & 49.0 & 48.8 & 49.0 & 50.7 & 46.6 & 48.3 & 49.1 & 49.0 & 49.0 & 48.8 \\ 
\gl--\en & 64.1 & 64.4 & 64.7 & 62.2 & 64.4 & 62.5 & 63.4 & 64.4 & 62.4 & 63.0 & 63.6 \\ 
\gl--\es & 71.3 & 71.5 & 71.5 & 72.1 & 71.7 & 71.1 & 71.0 & 71.6 & 71.4 & 72.5 & 71.6 \\ 
\gl--\pt & 66.9 & 67.1 & 67.4 & 67.6 & 67.5 & 67.7 & 67.1 & 67.6 & 66.8 & 68.1 & 67.4 \\ 
\gl--\ru & 46.7 & 46.5 & 45.9 & 45.0 & 46.3 & 42.8 & 45.8 & 44.8 & 44.7 & 45.7 & 45.4 \\ 
\gl--\sk & 48.2 & 48.1 & 47.2 & 48.5 & 48.8 & 45.3 & 47.6 & 46.7 & 45.5 & 48.2 & 47.4 \\ 
\gl--\tr & 39.7 & 39.3 & 39.3 & 39.1 & 38.2 & 35.9 & 38.8 & 38.9 & 38.3 & 38.0 & 38.5 \\ 
\pt--\az & 11.7 & 14.6 & 13.4 & 14.6 & 15.2 & 12.5 & 13.4 & 13.1 & 13.4 & 15.7 & 13.8 \\ 
\pt--\be & 18.9 & 17.2 & 18.2 & 21.0 & 18.7 & 20.2 & 18.7 & 19.7 & 18.4 & 18.7 & 19.0 \\ 
\pt--\cs & 71.6 & 72.0 & 70.6 & 71.7 & 71.7 & 72.0 & 71.5 & 71.9 & 71.2 & 70.7 & 71.5 \\ 
\pt--\en & 84.0 & 84.3 & 84.1 & 85.1 & 84.2 & 84.9 & 84.1 & 83.9 & 84.7 & 84.3 & 84.4 \\ 
\pt--\es & 92.8 & 93.2 & 93.2 & 93.2 & 93.6 & 93.0 & 93.4 & 93.3 & 93.2 & 93.4 & 93.2 \\ 
\pt--\gl & 49.3 & 49.6 & 48.9 & 50.1 & 49.9 & 46.8 & 49.3 & 48.9 & 47.9 & 49.6 & 49.0 \\ 
\pt--\ru & 63.6 & 64.3 & 62.8 & 64.7 & 64.4 & 64.3 & 63.0 & 63.4 & 63.8 & 62.4 & 63.7 \\ 
\pt--\sk & 63.6 & 62.4 & 62.6 & 63.9 & 63.0 & 62.6 & 62.4 & 62.1 & 59.7 & 62.2 & 62.4 \\ 
\pt--\tr & 60.4 & 60.8 & 60.4 & 62.3 & 59.5 & 60.4 & 60.3 & 60.9 & 60.5 & 60.9 & 60.6 \\ 
\ru--\az & 15.4 & 17.0 & 18.7 & 20.1 & 18.4 & 18.4 & 19.0 & 17.9 & 17.3 & 19.8 & 18.2 \\ 
\ru--\be & 25.1 & 22.2 & 24.5 & 23.8 & 24.3 & 24.0 & 24.5 & 24.3 & 25.3 & 24.3 & 24.2 \\ 
\ru--\cs & 70.8 & 70.3 & 70.9 & 70.4 & 70.8 & 71.3 & 71.0 & 70.5 & 70.8 & 71.1 & 70.8 \\ 
\ru--\en & 76.9 & 77.8 & 78.6 & 76.6 & 78.4 & 77.8 & 77.4 & 76.8 & 77.1 & 77.5 & 77.5 \\ 
\ru--\es & 75.2 & 75.2 & 75.3 & 76.3 & 75.6 & 75.3 & 76.3 & 74.8 & 76.4 & 74.5 & 75.5 \\ 
\ru--\gl & 43.1 & 42.2 & 42.1 & 43.3 & 43.5 & 37.1 & 41.9 & 41.7 & 41.3 & 40.5 & 41.7 \\ 
\ru--\pt & 72.6 & 71.8 & 72.6 & 74.5 & 72.5 & 72.6 & 71.5 & 71.5 & 72.2 & 70.2 & 72.2 \\ 
\ru--\sk & 65.5 & 66.8 & 66.3 & 66.5 & 66.3 & 66.4 & 67.0 & 66.5 & 64.7 & 66.9 & 66.3 \\ 
\ru--\tr & 56.1 & 56.2 & 55.2 & 57.7 & 56.8 & 57.0 & 56.1 & 54.8 & 57.3 & 54.8 & 56.2 \\ 
\sk--\az & 11.0 & 11.0 & 10.7 & 13.8 & 10.7 & 13.2 & 13.2 & 10.4 & 11.3 & 12.0 & 11.7 \\ 
\sk--\be & 23.2 & 20.8 & 21.1 & 22.1 & 21.1 & 22.9 & 22.7 & 22.9 & 23.4 & 22.1 & 22.2 \\ 
\sk--\cs & 85.1 & 85.5 & 84.6 & 84.4 & 85.3 & 85.9 & 85.6 & 84.9 & 85.0 & 85.0 & 85.1 \\ 
\sk--\en & 74.5 & 76.3 & 76.6 & 73.9 & 75.7 & 76.0 & 75.6 & 75.4 & 75.3 & 75.8 & 75.5 \\ 
\sk--\es & 75.7 & 75.5 & 74.9 & 76.2 & 74.4 & 74.2 & 74.6 & 74.4 & 74.7 & 74.7 & 74.9 \\ 
\sk--\gl & 49.1 & 48.7 & 48.9 & 51.7 & 50.1 & 40.9 & 49.4 & 48.5 & 49.6 & 49.7 & 48.7 \\ 
\sk--\pt & 73.7 & 73.2 & 72.6 & 74.7 & 74.0 & 73.1 & 71.7 & 72.8 & 72.9 & 72.0 & 73.1 \\ 
\sk--\ru & 63.5 & 64.4 & 62.8 & 64.0 & 64.0 & 64.2 & 64.0 & 62.6 & 62.6 & 64.6 & 63.7 \\ 
\sk--\tr & 55.4 & 57.0 & 56.2 & 57.4 & 55.7 & 55.4 & 57.0 & 56.0 & 54.4 & 55.2 & 56.0 \\ 
\tr--\az & 22.9 & 24.6 & 23.9 & 23.2 & 23.6 & 24.9 & 23.6 & 23.2 & 24.6 & 24.9 & 23.9 \\ 
\tr--\be & 22.2 & 16.8 & 21.6 & 20.7 & 21.3 & 21.6 & 23.4 & 19.8 & 19.5 & 21.3 & 20.8 \\ 
\tr--\cs & 68.5 & 68.0 & 66.7 & 67.2 & 68.0 & 68.1 & 68.4 & 67.1 & 67.8 & 66.3 & 67.6 \\ 
\tr--\en & 73.5 & 74.0 & 73.7 & 73.2 & 73.0 & 73.2 & 74.2 & 74.0 & 72.9 & 72.2 & 73.4 \\ 
\tr--\es & 74.4 & 74.0 & 74.6 & 75.5 & 73.2 & 73.8 & 74.6 & 74.7 & 74.8 & 74.4 & 74.4 \\ 
\tr--\gl & 36.1 & 36.6 & 35.9 & 36.4 & 35.9 & 29.7 & 36.7 & 36.7 & 35.0 & 36.8 & 35.6 \\ 
\tr--\pt & 72.2 & 71.8 & 71.8 & 72.8 & 71.3 & 71.4 & 70.8 & 71.8 & 72.4 & 72.1 & 71.8 \\ 
\tr--\ru & 61.3 & 61.8 & 60.0 & 61.8 & 61.7 & 61.8 & 60.5 & 60.0 & 59.5 & 59.9 & 60.8 \\ 
\tr--\sk & 55.4 & 56.8 & 56.8 & 57.0 & 56.2 & 54.9 & 56.4 & 55.8 & 51.6 & 55.4 & 55.6 \\ 
\bottomrule
    \end{tabular}
\end{table*}

\begin{table*}[t]
    \centering
    \caption{All results from the distant languages MWE experiment (P@1).}
    \label{tab:app3a}
    \begin{tabular}{c|ccccccc|c}
    \toprule
    \multirow{2}{*}{Test} & \multicolumn{7}{|c|}{Hub language} & \multirow{2}{*}{$\mu$} \\
     & \en & \fr & \hi & \ko & \ru & \sv & \uk &   \\ \midrule
\en--\fr & 75.1 & 75.3 & 75.2 & 75.8 & 76.3 & 75.5 & 75.4 & 75.5 \\ 
\en--\hi & 20.9 & 23.5 & 21.0 & 21.4 & 23.5 & 21.4 & 23.9 & 22.2 \\ 
\en--\ko & 9.2 & 10.4 & 9.1 & 9.8 & 9.8 & 10.1 & 10.0 & 9.8 \\ 
\en--\ru & 41.8 & 42.0 & 41.8 & 41.5 & 42.0 & 41.8 & 42.0 & 41.8 \\ 
\en--\sv & 57.0 & 57.5 & 59.0 & 56.6 & 57.8 & 57.6 & 58.4 & 57.7 \\ 
\en--\uk & 26.9 & 27.5 & 26.9 & 26.9 & 28.3 & 27.8 & 26.2 & 27.2 \\ 
\fr--\en & 72.5 & 72.0 & 71.6 & 72.7 & 72.9 & 73.4 & 74.0 & 72.7 \\ 
\fr--\hi & 18.7 & 16.0 & 14.8 & 17.3 & 19.0 & 17.8 & 17.5 & 17.3 \\ 
\fr--\ko & 6.9 & 6.7 & 5.8 & 5.5 & 5.8 & 7.5 & 6.0 & 6.3 \\ 
\fr--\ru & 39.9 & 38.3 & 40.3 & 40.4 & 40.8 & 40.0 & 39.6 & 39.9 \\ 
\fr--\sv & 51.8 & 49.3 & 50.5 & 51.1 & 49.4 & 48.2 & 51.8 & 50.3 \\ 
\fr--\uk & 28.8 & 27.0 & 27.8 & 28.5 & 28.7 & 27.7 & 26.1 & 27.8 \\ 
\hi--\en & 27.8 & 31.4 & 27.9 & 28.6 & 30.4 & 29.3 & 29.3 & 29.3 \\ 
\hi--\fr & 25.6 & 23.1 & 25.1 & 23.3 & 26.9 & 25.5 & 24.2 & 24.8 \\ 
\hi--\ko & 2.1 & 1.7 & 1.3 & 1.6 & 1.6 & 1.4 & 1.8 & 1.6 \\ 
\hi--\ru & 13.9 & 14.2 & 14.3 & 13.6 & 14.3 & 13.5 & 14.6 & 14.0 \\ 
\hi--\sv & 17.3 & 16.8 & 16.3 & 15.9 & 17.0 & 15.9 & 16.6 & 16.6 \\ 
\hi--\uk & 10.3 & 10.5 & 9.1 & 9.1 & 9.8 & 9.5 & 9.6 & 9.7 \\ 
\ko--\en & 15.1 & 16.6 & 15.2 & 17.0 & 16.6 & 17.7 & 16.4 & 16.4 \\ 
\ko--\fr & 11.9 & 10.2 & 10.9 & 10.9 & 12.6 & 13.6 & 10.8 & 11.6 \\ 
\ko--\hi & 1.8 & 2.4 & 1.2 & 1.6 & 2.0 & 1.8 & 2.0 & 1.9 \\ 
\ko--\ru & 7.9 & 6.6 & 6.0 & 5.7 & 6.9 & 6.8 & 7.3 & 6.7 \\ 
\ko--\sv & 6.8 & 6.6 & 5.9 & 5.9 & 7.2 & 5.6 & 7.2 & 6.5 \\ 
\ko--\uk & 3.5 & 3.6 & 3.4 & 3.2 & 3.5 & 3.5 & 3.1 & 3.4 \\ 
\ru--\en & 50.2 & 53.2 & 52.2 & 53.4 & 52.5 & 52.6 & 52.1 & 52.3 \\ 
\ru--\fr & 51.1 & 49.6 & 50.7 & 51.7 & 51.0 & 50.6 & 50.3 & 50.7 \\ 
\ru--\hi & 14.6 & 15.0 & 12.0 & 14.6 & 13.3 & 14.8 & 15.3 & 14.2 \\ 
\ru--\ko & 5.2 & 4.6 & 4.4 & 3.6 & 4.3 & 4.1 & 5.0 & 4.4 \\ 
\ru--\sv & 40.7 & 40.9 & 40.1 & 41.0 & 39.8 & 36.7 & 41.3 & 40.1 \\ 
\ru--\uk & 55.3 & 56.1 & 55.8 & 56.3 & 55.3 & 55.3 & 54.9 & 55.6 \\ 
\sv--\en & 51.2 & 51.1 & 52.3 & 51.9 & 52.0 & 50.7 & 52.7 & 51.7 \\ 
\sv--\fr & 47.9 & 45.7 & 46.8 & 48.2 & 47.1 & 46.6 & 47.4 & 47.1 \\ 
\sv--\hi & 17.2 & 16.3 & 15.0 & 16.0 & 17.7 & 15.9 & 17.0 & 16.4 \\ 
\sv--\ko & 4.9 & 4.2 & 4.0 & 3.8 & 5.0 & 4.0 & 5.1 & 4.4 \\ 
\sv--\ru & 31.5 & 33.2 & 32.4 & 33.0 & 31.8 & 30.2 & 31.8 & 32.0 \\ 
\sv--\uk & 22.4 & 23.8 & 23.0 & 23.5 & 24.1 & 21.0 & 21.9 & 22.8 \\ 
\uk--\en & 39.5 & 40.8 & 40.3 & 40.7 & 41.4 & 40.2 & 40.2 & 40.4 \\ 
\uk--\fr & 43.6 & 42.3 & 44.0 & 43.3 & 43.0 & 43.3 & 40.6 & 42.9 \\ 
\uk--\hi & 13.8 & 13.8 & 12.8 & 12.8 & 12.7 & 14.4 & 13.0 & 13.3 \\ 
\uk--\ko & 2.6 & 2.5 & 2.4 & 2.0 & 2.0 & 2.4 & 2.6 & 2.4 \\ 
\uk--\ru & 59.4 & 58.9 & 59.7 & 58.7 & 59.1 & 58.4 & 58.6 & 59.0 \\ 
\uk--\sv & 35.8 & 35.5 & 35.8 & 36.8 & 35.4 & 32.7 & 35.1 & 35.3 \\ 
    \bottomrule
    \end{tabular}
\end{table*}

\begin{table*}[t]
    \centering
    \caption{All results from the distant languages MWE experiment (P@5).}
    \label{tab:app3b}
    \begin{tabular}{c|ccccccc|c}
    \toprule
    \multirow{2}{*}{Test} & \multicolumn{7}{|c|}{Hub language} & \multirow{2}{*}{$\mu$} \\
     & \en & \fr & \hi & \ko & \ru & \sv & \uk &   \\ \midrule
\en--\fr & 87.3 & 88.2 & 87.8 & 88.4 & 88.3 & 88.0 & 87.7 & 88.0 \\ 
\en--\hi & 37.2 & 39.4 & 36.5 & 37.1 & 39.3 & 38.7 & 39.9 & 38.3 \\ 
\en--\ko & 23.4 & 24.6 & 22.6 & 23.4 & 24.3 & 25.9 & 25.0 & 24.2 \\ 
\en--\ru & 63.5 & 65.3 & 65.1 & 64.8 & 66.9 & 64.6 & 65.9 & 65.2 \\ 
\en--\sv & 74.8 & 76.1 & 76.3 & 75.8 & 75.4 & 75.6 & 76.5 & 75.8 \\ 
\en--\uk & 47.7 & 49.8 & 49.3 & 47.9 & 49.3 & 48.5 & 47.7 & 48.6 \\ 
\fr--\en & 85.3 & 84.5 & 83.7 & 84.5 & 85.4 & 85.1 & 84.6 & 84.7 \\ 
\fr--\hi & 32.7 & 30.0 & 29.5 & 30.6 & 33.4 & 32.2 & 31.6 & 31.4 \\ 
\fr--\ko & 14.9 & 14.5 & 14.0 & 14.6 & 16.0 & 15.3 & 15.2 & 14.9 \\ 
\fr--\ru & 61.0 & 59.5 & 61.9 & 61.7 & 62.1 & 60.6 & 60.9 & 61.1 \\ 
\fr--\sv & 69.6 & 68.1 & 68.8 & 69.1 & 68.6 & 68.0 & 71.1 & 69.0 \\ 
\fr--\uk & 45.6 & 44.2 & 44.8 & 45.6 & 45.8 & 45.0 & 44.1 & 45.0 \\ 
\hi--\en & 44.5 & 47.0 & 46.3 & 44.3 & 47.0 & 46.3 & 46.7 & 46.0 \\ 
\hi--\fr & 41.7 & 39.3 & 41.6 & 39.6 & 42.7 & 41.2 & 42.3 & 41.2 \\ 
\hi--\ko & 5.3 & 4.8 & 3.4 & 3.5 & 4.7 & 5.1 & 5.0 & 4.5 \\ 
\hi--\ru & 27.6 & 29.6 & 27.6 & 28.1 & 27.9 & 28.8 & 29.5 & 28.4 \\ 
\hi--\sv & 31.7 & 31.7 & 30.8 & 30.7 & 32.7 & 30.2 & 32.0 & 31.4 \\ 
\hi--\uk & 21.4 & 21.9 & 19.9 & 20.1 & 20.8 & 20.4 & 20.2 & 20.7 \\ 
\ko--\en & 28.9 & 28.7 & 27.0 & 28.1 & 30.1 & 33.1 & 28.6 & 29.2 \\ 
\ko--\fr & 21.9 & 21.6 & 19.7 & 20.4 & 24.0 & 24.4 & 21.3 & 21.9 \\ 
\ko--\hi & 4.3 & 4.8 & 3.9 & 4.1 & 4.6 & 4.8 & 5.0 & 4.5 \\ 
\ko--\ru & 16.2 & 15.3 & 12.9 & 13.4 & 15.8 & 15.7 & 16.3 & 15.1 \\ 
\ko--\sv & 16.2 & 14.1 & 13.9 & 13.8 & 15.6 & 13.9 & 16.3 & 14.8 \\ 
\ko--\uk & 9.7 & 8.0 & 8.6 & 8.6 & 9.3 & 8.2 & 8.8 & 8.8 \\ 
\ru--\en & 69.8 & 71.1 & 70.9 & 71.0 & 70.2 & 71.1 & 71.3 & 70.8 \\ 
\ru--\fr & 65.7 & 66.2 & 67.7 & 67.9 & 67.0 & 66.6 & 67.2 & 66.9 \\ 
\ru--\hi & 27.3 & 27.6 & 24.7 & 26.7 & 25.6 & 26.6 & 28.7 & 26.7 \\ 
\ru--\ko & 12.1 & 10.4 & 10.1 & 10.0 & 11.1 & 10.4 & 12.4 & 10.9 \\ 
\ru--\sv & 58.8 & 58.9 & 58.2 & 58.2 & 58.8 & 56.1 & 59.9 & 58.4 \\ 
\ru--\uk & 68.3 & 68.8 & 69.2 & 68.0 & 68.8 & 68.6 & 66.9 & 68.4 \\ 
\sv--\en & 65.4 & 66.2 & 66.3 & 65.7 & 65.1 & 64.4 & 65.9 & 65.6 \\ 
\sv--\fr & 62.5 & 60.1 & 60.3 & 61.1 & 60.7 & 59.8 & 61.3 & 60.8 \\ 
\sv--\hi & 28.2 & 28.0 & 26.6 & 27.4 & 29.3 & 27.1 & 28.6 & 27.9 \\ 
\sv--\ko & 11.7 & 10.7 & 10.9 & 9.8 & 11.5 & 11.6 & 11.4 & 11.1 \\ 
\sv--\ru & 50.5 & 51.0 & 50.7 & 50.9 & 50.3 & 47.8 & 49.9 & 50.2 \\ 
\sv--\uk & 40.2 & 42.1 & 41.6 & 41.6 & 41.7 & 38.3 & 39.2 & 40.6 \\ 
\uk--\en & 56.3 & 58.1 & 57.5 & 57.2 & 59.1 & 58.1 & 56.1 & 57.5 \\ 
\uk--\fr & 58.3 & 56.4 & 58.5 & 58.7 & 58.9 & 58.0 & 56.4 & 57.9 \\ 
\uk--\hi & 27.2 & 25.8 & 24.0 & 25.4 & 26.5 & 25.8 & 25.3 & 25.7 \\ 
\uk--\ko & 7.4 & 7.2 & 6.8 & 6.0 & 7.3 & 7.3 & 7.3 & 7.0 \\ 
\uk--\ru & 71.0 & 71.0 & 71.2 & 70.1 & 70.4 & 70.7 & 70.5 & 70.7 \\ 
\uk--\sv & 53.3 & 53.3 & 52.5 & 53.1 & 53.7 & 48.9 & 53.1 & 52.5 \\ 
    \bottomrule
    \end{tabular}
\end{table*}

\begin{table*}[t]
    \centering
    \caption{All results from the distant languages MWE experiment (P@10).}
    \label{tab:app3c}
    \begin{tabular}{c|ccccccc|c}
    \toprule
    \multirow{2}{*}{Test} & \multicolumn{7}{|c|}{Hub language} & \multirow{2}{*}{$\mu$} \\
     & \en & \fr & \hi & \ko & \ru & \sv & \uk &   \\ \midrule
\en--\fr & 90.8 & 91.3 & 90.1 & 91.0 & 91.1 & 91.1 & 90.7 & 90.9 \\ 
\en--\hi & 44.0 & 45.9 & 43.3 & 43.1 & 45.0 & 45.2 & 45.6 & 44.6 \\ 
\en--\ko & 31.1 & 31.5 & 28.4 & 30.5 & 31.6 & 33.7 & 32.1 & 31.3 \\ 
\en--\ru & 70.1 & 71.7 & 71.0 & 70.7 & 72.4 & 71.1 & 72.3 & 71.3 \\ 
\en--\sv & 80.0 & 81.1 & 80.9 & 80.4 & 80.8 & 80.4 & 81.2 & 80.7 \\ 
\en--\uk & 55.3 & 57.5 & 56.5 & 55.2 & 57.4 & 56.4 & 54.6 & 56.1 \\ 
\fr--\en & 87.6 & 87.8 & 86.6 & 87.7 & 88.0 & 87.9 & 88.0 & 87.6 \\ 
\fr--\hi & 39.1 & 35.3 & 35.5 & 36.5 & 38.6 & 38.1 & 38.5 & 37.4 \\ 
\fr--\ko & 20.1 & 18.4 & 18.4 & 19.6 & 20.3 & 19.4 & 19.7 & 19.4 \\ 
\fr--\ru & 67.1 & 65.9 & 68.1 & 67.5 & 66.8 & 66.8 & 67.4 & 67.1 \\ 
\fr--\sv & 74.4 & 73.3 & 74.2 & 74.8 & 73.3 & 73.3 & 75.5 & 74.1 \\ 
\fr--\uk & 51.7 & 49.7 & 51.3 & 51.8 & 52.0 & 51.2 & 49.9 & 51.1 \\ 
\hi--\en & 50.0 & 52.3 & 53.0 & 50.8 & 52.7 & 51.7 & 52.3 & 51.8 \\ 
\hi--\fr & 49.0 & 45.5 & 46.8 & 46.8 & 48.3 & 48.1 & 48.9 & 47.6 \\ 
\hi--\ko & 7.9 & 7.2 & 5.1 & 5.1 & 6.4 & 6.6 & 7.2 & 6.5 \\ 
\hi--\ru & 34.5 & 35.3 & 34.5 & 34.7 & 33.6 & 35.3 & 36.3 & 34.9 \\ 
\hi--\sv & 38.0 & 37.5 & 36.1 & 37.9 & 38.9 & 36.3 & 38.5 & 37.6 \\ 
\hi--\uk & 27.3 & 27.6 & 25.8 & 25.4 & 26.2 & 25.9 & 25.5 & 26.3 \\ 
\ko--\en & 34.2 & 34.3 & 32.3 & 35.2 & 37.1 & 38.4 & 35.4 & 35.3 \\ 
\ko--\fr & 27.0 & 25.9 & 23.7 & 24.6 & 28.5 & 30.1 & 26.4 & 26.6 \\ 
\ko--\hi & 6.2 & 6.9 & 5.6 & 6.0 & 6.7 & 6.7 & 6.9 & 6.4 \\ 
\ko--\ru & 21.2 & 19.3 & 16.4 & 18.2 & 20.4 & 20.9 & 20.8 & 19.6 \\ 
\ko--\sv & 20.9 & 18.1 & 17.8 & 17.5 & 21.1 & 18.4 & 20.6 & 19.2 \\ 
\ko--\uk & 12.9 & 12.1 & 11.5 & 11.3 & 12.6 & 12.0 & 11.7 & 12.0 \\ 
\ru--\en & 74.9 & 75.8 & 75.4 & 75.5 & 75.5 & 76.2 & 75.6 & 75.6 \\ 
\ru--\fr & 71.8 & 72.5 & 73.0 & 72.2 & 72.7 & 72.7 & 72.6 & 72.5 \\ 
\ru--\hi & 33.0 & 32.9 & 30.1 & 32.1 & 31.9 & 32.1 & 34.6 & 32.4 \\ 
\ru--\ko & 17.2 & 14.6 & 13.2 & 13.5 & 15.9 & 15.0 & 16.7 & 15.2 \\ 
\ru--\sv & 64.7 & 64.7 & 63.6 & 64.6 & 64.2 & 62.5 & 64.6 & 64.1 \\ 
\ru--\uk & 73.3 & 72.8 & 73.1 & 72.0 & 73.1 & 72.9 & 71.7 & 72.7 \\ 
\sv--\en & 69.5 & 70.4 & 71.0 & 70.6 & 70.9 & 69.3 & 70.0 & 70.2 \\ 
\sv--\fr & 67.0 & 64.2 & 65.0 & 65.3 & 65.5 & 64.2 & 65.7 & 65.3 \\ 
\sv--\hi & 33.6 & 32.6 & 32.0 & 30.9 & 33.3 & 31.9 & 33.2 & 32.5 \\ 
\sv--\ko & 15.7 & 14.7 & 14.0 & 12.9 & 15.7 & 14.9 & 15.6 & 14.8 \\ 
\sv--\ru & 57.2 & 56.4 & 56.5 & 56.2 & 56.4 & 53.8 & 56.4 & 56.1 \\ 
\sv--\uk & 47.5 & 47.9 & 47.7 & 47.7 & 48.5 & 44.8 & 46.4 & 47.2 \\ 
\uk--\en & 61.6 & 63.4 & 62.9 & 62.2 & 63.5 & 62.7 & 61.1 & 62.5 \\ 
\uk--\fr & 63.5 & 62.4 & 63.9 & 63.4 & 64.3 & 63.5 & 61.9 & 63.3 \\ 
\uk--\hi & 32.7 & 32.3 & 28.6 & 30.2 & 31.7 & 31.5 & 30.7 & 31.1 \\ 
\uk--\ko & 10.6 & 10.2 & 9.5 & 8.7 & 10.1 & 10.4 & 10.2 & 10.0 \\ 
\uk--\ru & 74.5 & 73.8 & 74.1 & 73.9 & 74.5 & 74.1 & 73.9 & 74.1 \\ 
\uk--\sv & 59.1 & 58.8 & 58.8 & 58.7 & 59.3 & 55.2 & 57.8 & 58.2 \\      
    \bottomrule
    \end{tabular}
\end{table*}

\end{document}